\documentclass[10pt,journal,compsoc]{IEEEtran}

\ifCLASSOPTIONcompsoc
\usepackage[nocompress]{cite}
\else
\usepackage{cite}
\fi
\ifCLASSINFOpdf
\else
\fi
\usepackage{amsmath,amssymb,amsfonts}
\usepackage{graphicx}
\usepackage{textcomp}
\usepackage{xcolor}

\usepackage{mathrsfs}

\usepackage{amsmath}
\usepackage{amsfonts}
\usepackage{booktabs}
\usepackage{amssymb}
\usepackage{multirow}
\usepackage[colorlinks,linkcolor=red,citecolor=blue]{hyperref}
\usepackage{algorithm}
\usepackage{algpseudocode}
\usepackage{graphicx}
\usepackage{amsthm}
\usepackage{subcaption}

\graphicspath{{figure/}}
\usepackage{CJK}
\usepackage{indentfirst}
\usepackage{cases}

\newtheorem{definition}{Definition}[section]

\usepackage{bm}
\usepackage{color,soul}

\usepackage{tikz}

\usepackage{multirow}
\usepackage{adjustbox}
\usepackage{environ}
\usepackage{setspace} 

\newenvironment{alignSmall}{\nobreak\small\noindent\align}{\endalign}
\newenvironment{alignFootnotesize}{\nobreak\footnotesize\noindent\align}{\endalign}
\newenvironment{alignScriptsize}{\nobreak\scriptsize\noindent\align}{\endalign}

\NewEnviron{myAlignS}{%
	\begin{singlespace}
		\vspace{-3pt}
		\begin{alignSmall}
			\BODY
		\end{alignSmall}
		\vspace{-5pt}
	\end{singlespace}
}

\NewEnviron{myAlignSS}{%
	\begin{singlespace}
		\vspace{-3pt}
		\begin{alignFootnotesize}
			\BODY
		\end{alignFootnotesize}
		\vspace{-5pt}
	\end{singlespace}
}

\NewEnviron{myAlignSSS}{%
	\begin{singlespace}
		\vspace{-5pt}
		\begin{alignScriptsize}
			\BODY
		\end{alignScriptsize}
		\vspace{-5pt}
	\end{singlespace}
}

\usepackage{enumitem}

\newcommand{\equref}[1]{Equation (\ref{#1})}

\newcommand{\secref}[1]{Section \ref{#1}}
\newcommand{\tableref}[1]{Table~\ref{#1}}
  
\newcommand{\figref}[1]{Figure \ref{#1}} 
\graphicspath{{image/}}
\begin{document}
	
	\title{ ST-ExpertNet:  A Deep  Expert Framework for Traffic Prediction
	}
	
	%
	%
	%
	%
	

	\author{Hongjun~Wang\IEEEauthorrefmark{1},
		Jiyuan~Chen\IEEEauthorrefmark{1},
		Zipei~Fan\IEEEauthorrefmark{2},
		Zhiwen~Zhang,
		Zekun~Cai,
		and Xuan Song\IEEEauthorrefmark{2}
		\IEEEcompsocitemizethanks{
			\IEEEcompsocthanksitem Hongjun Wang, Jiyuan Chen and Xuan Song are with the: (1) SUSTech-UTokyo Joint Research Center on Super Smart City, Department of Computer Science and Engineering, Southern University of Science and Technology (SUSTech), Shenzhen, China. (2) Research Institute of Trustworthy Autonomous Systems, Southern University of Science and Technology (SUSTech), Shenzhen, China. (3) Guangdong Provincial Key Laboratory of Brain-inspired Intelligent Computation, Department of Computer Science and Engineering, Southern University of Science and Technology, Shenzhen, China. E-mail: {wanghj2020,11811810}@mail.sustech.edu.cn and songx@sustech.edu.cn.   \hfil\break 
			\IEEEcompsocthanksitem Zipei Fan, Zhiwen Zhang and Zekun Cai are The University
			of Tokyo, 5-1-5 Kashiwanoha, Kashiwa-shi, Chiba, 277-8561, Japan; emails: fanzipei@iis.u-tokyo.ac.jp and {zhangzhiwen, caizekun}@csis.u-tokyo.ac.jp \hfil\break 
			\IEEEcompsocthanksitem  H.Wang and J.Chen contribute equally to this work
			\IEEEcompsocthanksitem  Corresponding to Z.Fan and X.Song
		}
	}

	%
	%

\markboth{Journal of \LaTeX\ Class Files,~Vol.~XX, No.~X, August~201X}%
{Shell \MakeLowercase{\textit{et al.}}: Bare Demo of IEEEtran.cls for Computer Society Journals}
%

\IEEEtitleabstractindextext{%
	\begin{abstract}
		Recently, forecasting the crowd flows has become an important research topic, and plentiful technologies have achieved good performances. As we all know, the flow at a citywide level is in a mixed state with several basic patterns (e.g., commuting, working, and commercial) caused by the city area functional distributions (e.g., developed commercial areas, educational areas and parks). However, existing technologies have been criticized for their lack of considering the differences in the flow patterns among regions since they want to build only one comprehensive model to learn the mixed flow tensors. Recognizing this limitation, we present a new perspective on flow prediction and propose an explainable framework named ST-ExpertNet, which can adopt every spatial-temporal model and train a set of functional experts devoted to specific flow patterns. Technically, we train a bunch of experts based on the Mixture of Experts (MoE), which guides each expert to specialize in different kinds of flow patterns in sample spaces by using the gating network. We define several criteria, including comprehensiveness, sparsity, and preciseness, to construct the experts for better interpretability and performances. We conduct experiments on a wide range of real-world taxi and bike datasets in Beijing and NYC. The visualizations of the expert's intermediate results demonstrate that our ST-ExpertNet successfully disentangles the city's mixed flow tensors along with the city layout, e.g., the urban ring road structure. Different network architectures, such as ST-ResNet, ConvLSTM, and CNN, have been adopted into our ST-ExpertNet framework for experiments and the results demonstrates the superiority of our framework in both interpretability and  performances.
		
	\end{abstract}
	
	\begin{IEEEkeywords}
		Crowd Flow Prediction, Mixture of Experts, Neural Networks, Urban Computing
\end{IEEEkeywords}}

\maketitle

\section{Introduction}
\IEEEPARstart{I}n recent years, with the maturity of perception technology and computing environment, various kinds of big data have quietly emerged in cities, such as traffic flow, weather data, road networks, points of interest, movement trajectories, and social media. Among them, the task of traffic flow prediction is of great significance to urban traffic management and public safety and has received extensive attention from academia in recent years. Existing efforts have mainly considered that traffic flow is affected by recent time intervals, daily periodicity, and weekly trend \cite{zhang2017deep,yao2019revisiting,lin2019deepstn,zonoozi2018periodic}, which can help the model distinguish functional regions. For instance, we can recognize between
residential areas and office areas by recent traffic flows. Specifically, people go out to work on weekdays morning, so the outflow of residential regions will increase, and the inflow will grow in office areas. Furthermore, by observing periodic flows, we can find that both inflow and outflow remain stable during the week in the residential area, but will drop significantly on weekends in the office area.

However, in the literature, these factors are mainly considered as purely time-series signals rather than the unique features of each functional region. Moreover, they ignored the fact that the flows of the different functional areas only influence relevant parts rather than citywide. For instance, the entertaining region's flow growing up on weekends will not affect the flow state in the office areas. This ignorance will finally result in the improper design of the previous citywide prediction model, which aims to use a single network to predict the flows of all functional regions, inevitably causing difficulties in thoroughly understanding the different flow patterns. Further, when an abnormal prediction is produced from the previous model, it can't tell why this happens. This will be solved if we can see the knowledge of the area learned by the model.
For instance, if the region is an entertaining district, the model considers it an office area and outputs tiny values on weekends. We will know if the misunderstanding of the model causes this.


Most of the previous work has successfully exploited the laws of human mobility in the flow tensor and provided a new perspective for researchers to understand the human flow pattern \cite{yuan2012discovering,yin2011geographical,wang2014discovering}. Recently, the crowd flow tensor has been demonstrated can be decomposed into several basic patterns of human life \cite{fan2014cityspectrum}, each of which has its law, such as the commuting pattern that focuses mainly on the morning and evening peaks on weekdays and the entertaining pattern that focuses on the entire day of the weekends. The flow pattern distribution is greatly affected by both time and the distribution of functional regions geographically. For example, the commuting pattern dominates highways all day while the working pattern only occupies office area till evenings. A functional region can be filled with many flow patterns at each timestamp, but only one or two of them is the main component.



\begin{figure}[t]
	\centering
	\includegraphics[width=1\linewidth]{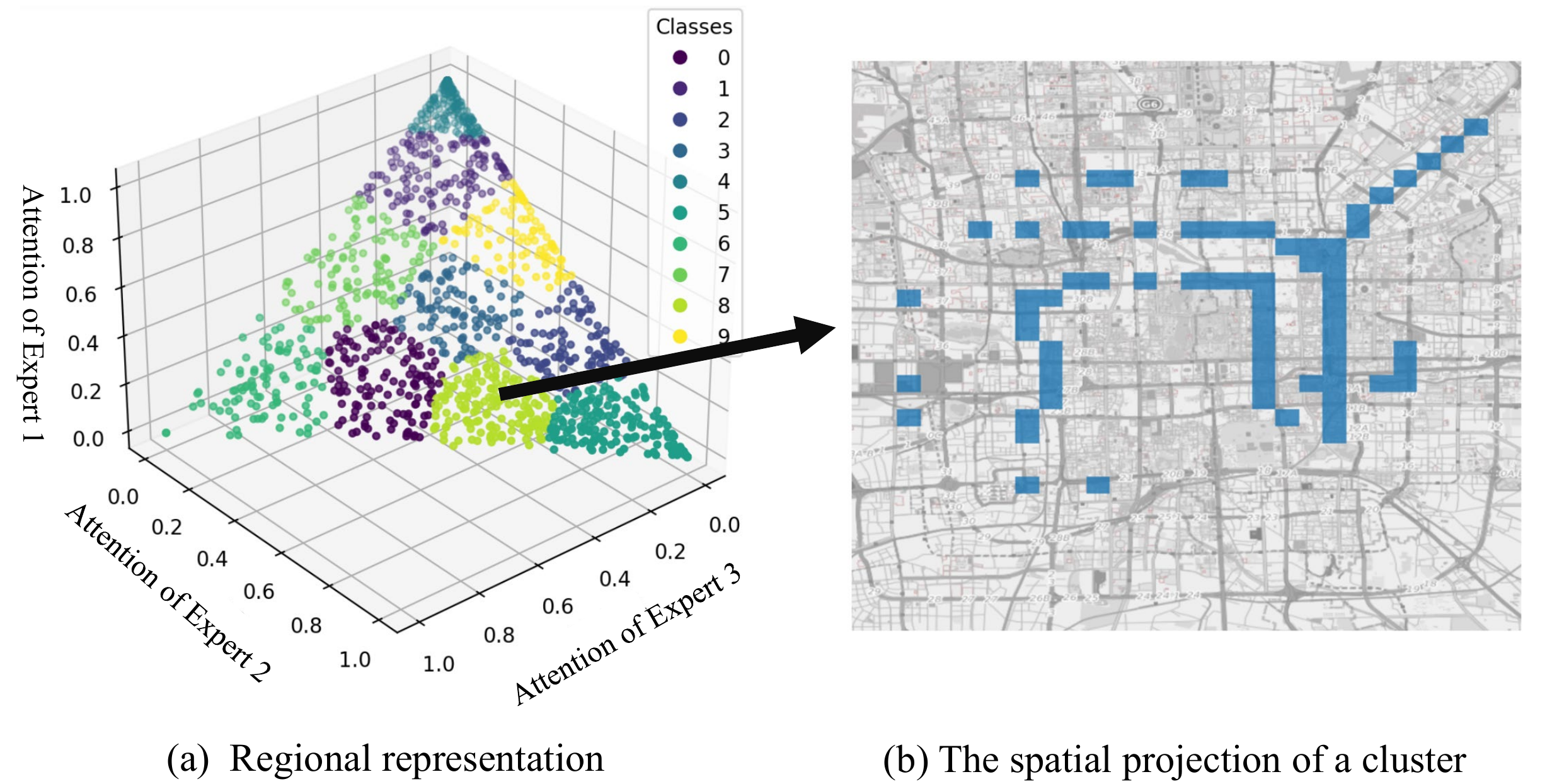}
	\caption{The relatively proportional weight of each expert to a certain region is shown on (a) and one of the clustered classes of  KMeans is depicted on (b). It is obvious that this class concentrates on the commuting flow pattern of the 2nd and 3rd ring roads in Beijing during rush hours. }
	\label{fig:regionrepresent}
\end{figure}

\figref{fig:regionrepresent} shows an example of using three same networks (experts) trained with attention mechanism on real-world taxi data in Beijing \cite{zhang2017deep} during rush hours. The attention mechanism assigns weights to the experts so that each region of the $32\times32$ mesh grids (i.e., the Beijing city map) might receive predictions from a number of experts but with different weightings. In \figref{fig:regionrepresent} (a), there are $32\times32$ points representing all regions of Beijing, and each region receives attention from three experts, whose relatively proportional weight can be viewed from the axis. We further cluster regions with similar relatively proportional weights of three experts using KMeans, and one of the classes has been represented in \figref{fig:regionrepresent} (b), which is the primary commuting road section in Beijing, demonstrating that the experts have successfully mined the commuting flow patterns and the underneath Beijing ring road structure. However, as we can see in the \figref{fig:regionrepresent} (a), the flow patterns are closely connected to each other, which suggests that a single attention mechanism without regularization can not fulfill our needs.

By summing up the conclusions proposed by previous work, this paper considers that the traffic flow prediction task could be divided into two-part: 1) building a set of expert models to automatically disentangle the mixed-flow tensors with several inherent patterns. 2) predicting the future flow state of each pattern. The task of training multiple functional experts could be considered as the MoE (mixture of experts) problem  \cite{jacobs1991task,jacobs1997bias,jordan1994hierarchical}, which generates expert systems through differentiated learning strategies to improve the precision and robustness of the deep neural network.
We propose the ST-ExpertNet framework based on MoE \cite{jacobs1991adaptive}, which divides the sample space by a gating network. 
In ST-ExpertNet, the flow of a region is equal to the combination of multiple basic flow patterns, which enhances the robustness and certainty of the model compared to a single model. For example, we can see the reasonableness of the model's predictions through the components of different flow patterns. Meanwhile, as mentioned above, the combination of multiple models improves the fault tolerance of the model compared to a traditional model, which is a kind of approach widely used in adversarial robustness\cite{pang2019improving}.


For better performance and
interpretability of ST-ExpertNet, this paper considers the following criteria:

\textbf{\textit{Comprehensiveness}:} The data distributions learned by each expert should be as different as possible. To achieve that, each expert should have enough knowledge to distinguish different types of flow patterns caused by time and the characteristic distribution of functional regions. In our approach, the factors of hour, day, and week and metadata have been concatenated together to provide each expert the distinctive features for each region and flow patterns. This is different from the traditional strategy \cite{zhang2017deep,zhang2019flow,yao2018deep}. 


\textbf{\textit{Sparsity}:} The expert should be discrepant in both spatial and temporal aspects. For instance, in terms of the temporal aspects, while one expert is in charge of the peaking hours in residential areas, the other one is responsible for the midnight periods. From spatial aspects, the experts should pay attention to different functional regions. Therefore, for each input, only a few “activated” experts are encouraged. In this paper, the penalty term \textit{Expert Inter Discrepancy Loss} which bases on Determinant Point Process (DPP) has been proposed to 
punish the similarity between experts. Additionally, redundant experts should be avoided, as they
add to the complexity of the explanation but do not add additional
performance. Therefore, the number of experts should be controlled in small numbers to provide a clear explanation for users in ST-ExpertNet.

\textbf{\textit{Preciseness}:} 
Each expert should assume full responsibility for its prediction. Therefore, \textit{Expert Responsibility Loss} has been proposed to improve the prediction performance in the flow patterns charged by experts. Two types of gating networks have been proposed in terms of spatial and temporal aspects. The former intends to 
generate the attention for each expert and
to provide the opportunity to peek inside the experts by attention mechanisms \cite{bahdanau2014neural}, and the latter aims to adjust the output signal amplitude from a global perspective. 

To further evaluate the interpretability and performance of the ST-ExpertNet framework, we
integrate ST-ResNet \cite{zhang2017deep}, convolutional LSTM (ConvLSTM) \cite{lin2019deepstn}, and convolutional neural networks (CNN), which are popular baselines in the flow prediction task, into our framework. We test our explainable framework in real-world taxi and bike datasets: TaxiBJ, TaxiNYC, BikeNYC-I, and BikeNYC-II in Beijing and New York respectively. ST-ExpertNet can give embedded representations for each region or pattern in our city with spatial and temporal aspects. For temporality, we show the evolutionary process between the office and residential areas of Beijing. For spatiality, we show the knowledge distribution of each expert at the citywide level, which is well-aligned with the human cognition of our city.

We summarize our contributions to the field as follows: 
\begin{itemize}
	\item[1)] We provide a new perspective of viewing the crowd flow prediction problem by decomposing the mixed flow pattern into several basic ones and training "specialized" experts to gain better performances. We propose a novel and efficient architecture based on the mixture of experts (MoE), which can adopt every ST-model by making it as an expert.
	\item[2)]  We design \textit{Experts Inter Discrepancy Loss} and  \textit{ Experts Responsibility Loss } for regularizing the expert learning paradigm.
	\item[3)]  We conduct experiments on several real-world traffic datasets. The popular baselines in flow prediction: ST-ResNet, ConvLSTM, CNN have been employed in our framework, and the results show that our model is consistently better and more explainable than other state-of-the-art methods.
\end{itemize}

The remainder of our paper is arranged as follows: Section \ref{sec:related} introduces the related work, including mixture of experts, crowd flow prediction and data mining in human mobility. Section \ref{sec:Formulation} gives the primary definition of flows and the flow prediction problem. It also describes our basic idea to solve crowd flow prediction based on the general framework of mixture of experts. Section  \ref{sec:Methodology} introduces our ST-ExpertNet framework in detail. Section \ref{sec:exps} conducts both the quantitative and qualitative experiments on our ST-ExpertNet.
Section \ref{sec:concluds} discusses the conclusion and future works.

\section{ Related work}\label{sec:related}
In this section, we will discuss some relevant works about mixture of expert system, traffic flow prediction, as well as data mining in human mobility.
\subsection{Mixture of Experts}
Mixture of experts (MoE) belongs to a type of combining method, which is similar to boosting and negative correlation learning methods. The basic ideas of MoE is to train multiple experts, and each expert is assigned to a different part of the subtask. Therefore, MoE is a good compromise between a single global model or multiple local models, but one of the most important questions for MoE is how to divide a dataset into different parts. Now, MoE can be roughly divided according to whether the data is divided in advance \cite{masoudnia2014mixture}. Based on the implicit problem space partitioning \cite{tang2002input,gutta2000mixture,goodband2006mixture}, the prior information is used to partition the dataset by clustering methods. The key issues for these technologies are the reliability of the prior information and how to balance the data after clustering. To overcome this challenge,  a gating approach has been proposed to balance the learning strategy by guiding the experts to learn to handle different subsets of training data \cite{jacobs1991task,hansen1999combining,ubeyli2010differentiation}.
There are other kinds of work focuing on adopting different models such as Support Vector Machines (SVMs) \cite{collobert2002parallel}, Gaussian Processes \cite{tresp2001mixtures} and Dirichlet Processes \cite{shahbaba2009nonlinear}, or different expert configurations such as a hierarchical
structure \cite{yao2009hierarchical}, infinite numbers of experts \cite{rasmussen2002infinite}, and
adding experts sequentially \cite{aljundi2017expert} into the framework.
Our work extends the application of MoE systems into the framework of flow prediction in urban computing by dividing the flow tensor space. Further, our design serves as a more explainable and efficient framework which can adopt every ST-models, which aims to improve their performance and interpretation.

\subsection{Traffic Flow Prediction}
Recently, machine learning is increasingly being used in intelligent transportation system, such as, ordinal classification \cite{yildirim2019eboc}, intrusion detection \cite{bangui2021recent} and classify vehicle images \cite{wen2015rapid}. Among the literature, 
Traffic flow prediction has become the hottest topic in transportation. Traffic flow owns the characteristics of highly nonlinear time correlation and uncertainty \cite{deri2016big,zhan2016citywide}, which brought huge challenges to traffic flow prediction. The basic assumption is to consider the traffic flow prediction as a time series problem. Among them, the auto-regressive moving average model (ARMA) \cite{said1984testing} as the fundamental time series prediction model has been widely used in traffic flow. The variant methods based on the wavelet neural network \cite{cheng2017traffic} and difference integration \cite{box1970distribution} have been proposed. However, these kinds of models lack the necessary modeling of spatial dependence \cite{zhang2017deep}. To both consider the spatial and temporal factors, the deep learning model is first proposed \cite{zhang2016dnn} by combining the convolutional neural network (CNN) and Deep Neural Networks (DNN). The time series concepts of the hour, day, and week patterns are further defined \cite{zhang2017deep} and the residual network (ResNet) \cite{he2016deep} is used to process them respectively. A multitask adversarial ST-network is later proposed to infer the 
in-out flows and origin-destination (OD) simultaneously \cite{wang2020multi}. A deep irregular convolutional residual LSTM
network is proposed \cite{du2019deep} to model the challenges, such as, hybrid transportation lines, mixed traffic, transfer stations,
and some extreme weathers.
Multiple architectures have been designed to improve the performance based on the framework \cite{zhang2017deep}, such as using ConvGRU \cite{zonoozi2018periodic}, Long Short-Term Memory (LSTM) \cite{yao2018deep}, ConvLSTM \cite{lin2019deepstn} instead of CNN.
Although these models achieved good performances, the criticisms are mainly focused on their lack of considering the discrimination of the flow patterns and try to use a single model to monitor them all. In this paper, we model the flow prediction problem from
a new perspective and intend to design a set of experts specialized in different kinds of flow patterns.


\subsection{Data Mining in Human Mobility}
Recently, a brunch of works focus on mining the inherent patterns from large-scale human mobility datasets. A series of approaches have been proposed for geographical topics  (like beach, hiking, and sunset) \cite{yin2011geographical}, city functional region (e.g., areas of historic interests, diplomatic and embassy areas) \cite{yuan2012discovering}, human trip purpose  (e.g., shopping, work, eating out) \cite{zhao2020discover}, space-time structure \cite{pozdnoukhov2011space}. Other part of works concentrates on using tensor factorization to explore the urban mobility patterns. The Non-negative Tensor Factorization (NTF) is utilized \cite{fan2014cityspectrum} to decompose the human flow tensor into several basic life patterns. The probabilistic tensor factorization framework \cite{sun2016understanding} is introduced to discover the travel behavior and spatial configuration of the city in Singapore.  A regularized non-negative Tucker decomposition
approach is applied \cite{wang2014discovering} to reveal the principal patterns from traffic networks evolving over time. These researches demonstrated the mixed state of human mobility and provide good theoretical motivation for using the MoE method, which is skilled in the handling of multiple sources of data.  To the best of our knowledge, we are the first to extend the idea of decomposing the mixed-state flow tensor into the flow prediction.

\section{Problem Formulation}\label{sec:Formulation}
As section 2.3 described, crowd flow at a citywide level is in a mixed state consisting of several basic patterns (e.g. commuting, working and commercial). Therefore, a natural thought would be to disentangle the patterns and study them separately before merging them to give the final prediction. This section mainly describes some primary definitions of flows and flow prediction problems, as well as our idea to disentangle the mixed flow pattern into basic ones using the mixture of experts structure. Note that the detailed implementation of our ST-ExpertNet framework is in Section \ref{sec:Methodology} and here we only express the idea of disentangling in an abstract manner.

\begin{definition}\label{def:region}
	\textbf{Region \cite{zhang2016dnn}}.
	A city is partitioned into $h\times w$ equal-size grids and each grid is called a \textit{region}. We use $(i,j)$ to represent a region that lies at the $i^{th}$ row and $j^{th}$ column of the city. \end{definition}


\begin{figure}[t]
	\centering
	\includegraphics[width=1\linewidth]{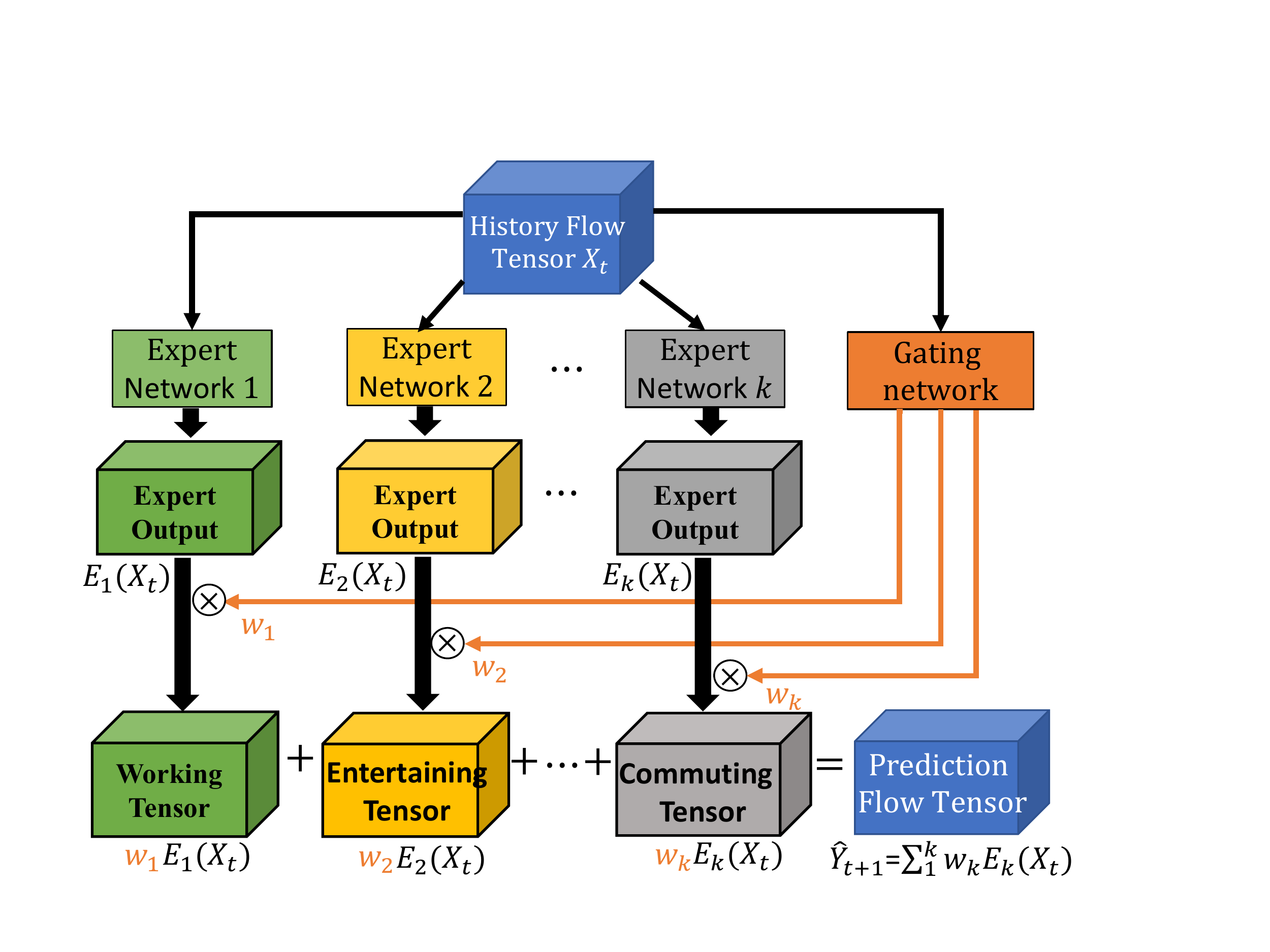}
	\caption{Schematic of our basic idea. We intend to build a set of models and a gating network, which can automatically disentangle the feature pattern (e.g. working, entertaining, commuting and commercial) from input and predict the future flow prediction by the weighted summation of each part. }
	\label{fig:formulation}
\end{figure}

\begin{definition}\label{def:flow}
	\textbf{Inflow and outflow of regions \cite{zhang2016dnn}}.
	Let $\mathbb{P}$ be the set of trajectories at the $t^{th}$ time interval. For a region $(i,j)$, the inflow and outflow at time interval $t$ can be defined as 
	
	\begin{align*}
		g_{t}^{i n, i, j} &=\sum_{T r \in \mathbb{P}}\left|\left\{k>1 \mid v_{k-1} \notin(i, j) \wedge v_{k} \in(i, j)\right\}\right| \\
		g_{t}^{o u t, i, j} &=\sum_{T r \in \mathbb{P}}\left|\left\{k \geq 1 \mid v_{k} \in(i, j) \wedge v_{k+1} \notin(i, j)\right\}\right|
	\end{align*}
	where, $T r: v_{1} \rightarrow v_{2} \rightarrow \cdots \rightarrow v_{|T r|}$ is a single trajectory in $\mathbb{P}$;
	$v_{k}\in(i, j)$ is a geospatial coordinate within a region $(i, j)$ and $| \cdot |$  denotes cardinality of a set.
\end{definition}


\begin{definition}\label{def:flowpred}
	\textbf{Flow prediction}.
	Given the historical observations 
	$X_t=\left\{g_{t^{\prime}} \mid t^{\prime} \in[0, n-1]\right\}$
	, predict future flow $\hat{Y}_{t+1}$.
	\begin{equation}
		\Phi: X_t \rightarrow \hat{Y}_{t+1}\label{equ:flow_pred}
	\end{equation}
	where $\Phi$ as neural network and $n$ denotes the length of the input time interval, $X_t \in \mathbb{R}^{2n \times h \times w}$ and $h, w$ denote the height and width of the city of grids. \\
\end{definition}

The Mixture of Experts (MoE) is a combination technique suitable for handling questions when the problem can be divided into a number of subtasks. Different from general neural networks, MoE separates and trains multiple expert models corresponding to each sub-task, and a gating module is used to assign weights to experts indicating their importances. The actual output of the model is the combination of the output of each expert, multiplying the weight from the gating module. 


\begin{table}[t]
	
	\centering
	\renewcommand\arraystretch{1.5}
	\caption{Symbol Description.}
	\label{tab:symbol}
	\small
	\begin{tabular}{c|c}
		\toprule
		Notation & Description\\
		\hline
		$g_{t}^{in}, g_{t}^{out}$ & Inflow and outflow matrix at $t^{th}$ time interval.\\
		$Y_{t+1}$ & Ground truth flow tensor at $(t+1)^{th}$ time interval.\\
		$\hat{Y}_{t+1}$ & The predicted flow tensor at $(t+1)^{th}$ time interval.\\
		$X_t^{ex}$ & The tensor of external information\\
		$X_t^{tr}$ & The tensor of trend (week) information\\
		$X_t^{p}$ & The tensor of period (day) information\\
		$X_t^{c}$ & The tensor of closeness (hour) information\\
		$X_t$ & The input tensor for each expert \\
		$K$ & The number of experts\\
		$G_s, G_t$ & The spatial and temporal gating network\\
		$E_i$ & The $i^{th}$ expert network\\
		\bottomrule
	\end{tabular}
\end{table}

In this case, we apply the idea of MoE to our question and feed the history flow tensor $X_t$ to all the experts and the gating network simultaneously. The gating network will then assign weights to each expert, helping to ensure its localization at a certain flow pattern. As a result, the mixed crowd flow tensors are automatically divided and charged by different experts.
\figref{fig:formulation} presents the idea with a graph. Formally, let the history flow tensor $X_t=\left\{g_{t^{\prime}} \mid t^{\prime} \in[t-n+1, t]\right\}$ and the set of expert models $\Phi=\{ E_1, E_2, \dots ,E_k\}$, the formulation of the output can be written as:
\begin{align}
	\hat{Y}_{t+1}= \sum_{E_i \in \Phi} W_i\odot E_i(X_t),\label{equ:prediction}
\end{align}
where $\odot$ denotes an element-wise product, $\hat{Y}_{t+1}$ denotes the future prediction flow tensor, $W_i \in \mathbb{R}^{2\times h \times w}$ is the weight of each model $E_i$ assigned by the gating network. Therefore, let ${Y}_{t+1}$ denote the ground-true future flow tensor, our first optimization goal can be written as:
\begin{align}
	\min || {Y}_{t+1}-  \sum_{E_i \in \Phi} W_i\odot E_i(X_t) ||^2.\label{equ:final_goal}
\end{align}

	

Since different flow patterns acts differently, similar output for each expert is unwanted. For instance, the working pattern focuses on the peak of morning and evening on weekdays, but the entertaining pattern mostly concentrated on the weekends.  
Therefore, to force the knowledge learned by each expert to be different, this paper tries to maximize the discrimination for each model's final output $W_i E_i(X_t)$, which can be formulated as
\begin{small}
	\begin{align}
		\max \sum_{E_i \in \Phi} \sum_{E_j \in \Phi} D_{eid}(W_i E_i(X_t),\ W_j E_j(X_t)),  ~E_i \neq E_j, ~\forall E_i,E_j, \label{equ:lambda1}
	\end{align}
\end{small}
where $D_{eid}$ denotes the distance metric between a pairwise model's final output. Moreover, to improve the interpretability and robustness of the experts, we hope each expert can be “responsible” for its own target pattern. The scenario is unwelcome when the gating network assigns the entertaining pattern to an expert but the expert itself tries to learn from the commuting pattern. We try to ensure this consistency between experts and the gating network by optimizing the following target which can be formulated as:
\begin{align}
	\min \sum_{E_i \in \Phi} W_i D_{er}({Y}_{t+1} -  E_i(X_t)) ,\label{equ:lambda2}
\end{align}
where $D_{er}$ denotes the distance between expert output and the future ground truth flow.

To sum up, our final optimization goals are threefold and described as Equation (\ref{equ:final_goal}), (\ref{equ:lambda1}) and (\ref{equ:lambda2}). The detailed implementation of the optimization process is demonstrated in the next section.



\section{Methodology}\label{sec:Methodology}

\begin{figure}[t]
	\centering
	\includegraphics[width=1\linewidth]{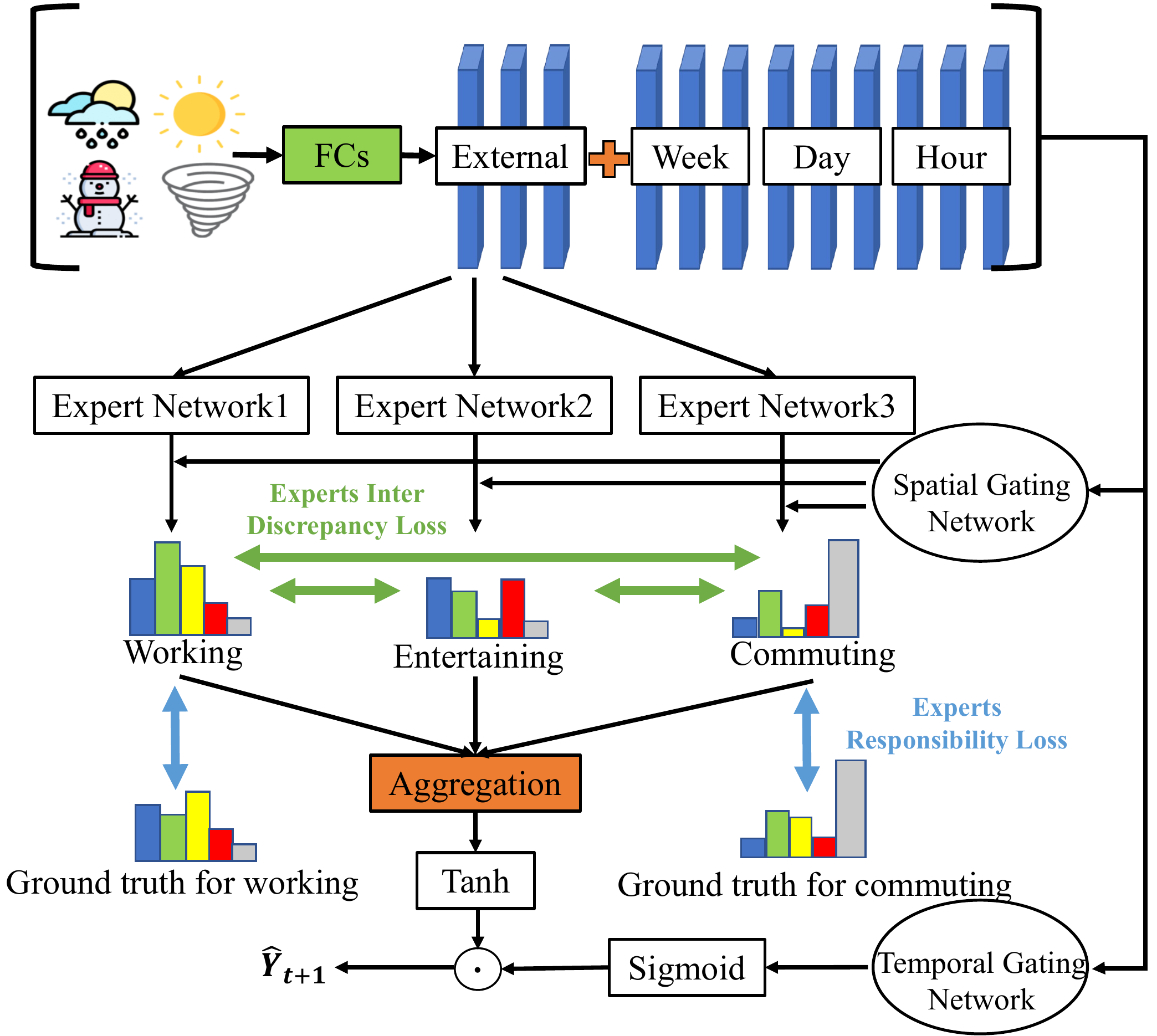}
	\caption{The framework of our ST-ExpertNet.}
	\label{fig:model}
\end{figure}
In this section, we introduce the architecture of ST-ExpertNet,  which is the framework based on MoE and aims to solve the optimization target \equref{equ:final_goal}, \equref{equ:lambda1} and \equref{equ:lambda2} in \secref{sec:Formulation}. 


\subsection{Overview}
We here systematically summarized our ST-ExperNet framework shown in \figref{fig:model}. The input of the expert network and gating network is composed of four parts: the historical observation of week, day, hour, and external information, which indicates the recent time intervals, daily periodicity,  weekly trend, and metadata (e.g., weather
conditions and events) respectively. While different expert networks produce their output, the spatial gating network further localizes them to their responsible patterns via the attention mechanism. The localized outputs of each expert are then aggregated together before the nonlinear transformation of \textit{Tanh}. Inspired by PixelCNN \cite{oord2016conditional}, which uses a gated activation unit to control the magnitude of the output, we introduce the temporal gating network to do the same thing. The output of the temporal gating network will pass through a \textit{Sigmoid} function before doing an element-wise multiplication with the aggregated experts' output. There are two regularization strategies shown in the figure: \textit{Expert Responsibility Loss} and \textit{Expert Inter Discrepancy Loss}. The former intends to force the experts to take responsibility for their sub-tasks by calculating the loss between their sub-task prediction (e.g., the prediction of commuting pattern) and the ground truth of the sub-task, and the latter aims to reduce the overlap of the responsible sub-tasks for every pair of experts. We organize the framework description as follows: The ST-ExpertNet input is introduced in \secref{sec:input}. The detailed implementation of ST-ExpertNet and its regularization are shown in \secref{st-expert} and \secref{regularization} respectively. Finally, we describe the training steps of our framework in \secref{training_step}.

\subsection{Multi-Source Information Fusion}\label{sec:input}

\figref{fig:model} shows the input of ST-ExpertNet, which
consists of four main components that we adopt from \cite{zhang2017deep}:\textit{ closeness, period, trend, and external influence} denoted as $ X_t^c, X_t^p, X_t^{tr}, X_t^{ex}$ respectively. However, unlike traditional operations in \cite{zhang2017deep,zonoozi2018periodic,lin2019deepstn,yao2019revisiting}, which feed these components into different models and hinder a comprehensive understanding of property of the 
flow patterns, our goal is to maximize the discrepancies between different types of flow patterns and minimize the differences between the same type (Equation \ref{equ:lambda1}), which can help experts to easily identify the regional inter-discrepancy (i.e., if regions (1,3) and (2,4) are highly relevant in the functional region of commuting flow pattern / commuting functional region). Given the set of functional region $\Omega=\{\Omega_1,\Omega_2, \cdots, \Omega_n\}$, where each region $R_i \in \Omega_s$ and $\Omega_s \in \Omega$ presents a type of functional region, e.g., office area. Our goal can be formulized with
\begin{small}
	\begin{align}
		\max	\sum_{\Omega_s \in \Omega}\left(\sum_{(R_i,\ R_j)\in \Omega_s} I(R_i;\ R_j) - \sum_{R_u \in \Omega_s,\ R_v \notin \Omega_s} I(R_u;\ R_v) \right),\label{equ:mutual}
	\end{align}
\end{small}
where $I$ denotes the mutual information between a pairwise of regions. Intuitively, the motivation of \equref{equ:mutual} is to minimize the distance between those regions in the same functional region and maximize the discrepancy between the regions in different functional regions. To achieve this goal, we concatenate \textit{  closeness, period, trend, and external} as an entirety and feed it into the experts and gating networks, which can give a comprehensive and distinct feature space for each region. The FC layer shown in \figref{fig:model} denotes the multilayer perceptron layer, which intends to process external information (e.g., weather condition). After reshaping the  FC's output as $X^{ex}_{t}\in \mathbb{R}^{n_{w} \times h\times w}$, where $n_w$ is the number of weather information, such as sunny, rainy, cloudy, we have the final ST-ExpertNet input $X_t$ as 
\begin{align*}
	&X_t^{tr}=\left\{g_{t^{\prime}} \mid t^{\prime} \in[t-(n_q+1)\times q, t-q]\right\}\\
	&X_t^{p}=\left\{g_{t^{\prime}} \mid t^{\prime} \in[t-(n_p+1)\times p, t-p]\right\}\\
	&X_t^{c}=\left\{g_{t^{\prime}} \mid t^{\prime} \in[t-(n_c+1), t]\right\}\\
	&X_t=Concat([X^{ex}_{t},\ X^{tr}_{t},\ X^p_{t},\ X^c_{t}])
\end{align*}where $n_q,n_p,n_c$ represent the length of time slots of Trend, Period, and Closeness, and $q, p$ are the length of one week and day respectively, for example, suppose $\Delta t=30$min, thus, $q=24h/0.5h=48$ and $p=q\times7$. Here $Concat$ denotes the channel-wise concatenate operator.


\subsection{ST-ExpertNet}\label{st-expert}
The structure of our ST-ExpertNet consists mainly of three parts: the Expert Layer, Spatial Gating Network, and Temporal Gating Network. In the following part, we will introduce them one by one.

\noindent\textbf{The Expert Layer.}
According to \secref{sec:Formulation}, our goal is to build a set of expert models, where each expert can take charge of a distinctive flow pattern within the flow tensor. This without any prior consultation holds the same view as the motivation of Mixture of Experts (MoE), which involves decomposing predictive modeling tasks into several subtasks, training the expert models corresponding to each subtask, developing a gating model that learns which expert to trust based on the input variables, and combining all the predictions from each expert together. The Original Mixture-of-Experts (MoE)  \cite{jacobs1991adaptive} can be formulated as:
\begin{align}
	\hat{y}=\sum_{i=1}^{n} G(x)_{i}\odot E_{i}(x),\label{equ:output}
\end{align}
where $x$ denotes the input of MoE and its size is associated with gating network and expert network, $\sum_{i=1}^{n} G(x)_{i}=1$ and $G(x)_{i}$, representing the $i^{th}$ logit of the output of $G(x)$, indicates the proportional contribution for expert $E_{i}$.
Here, $E_{i}, i=1, \ldots, n$ are the $n$ expert networks and $G$ represents a gating network that helps to group the results of all experts. More specifically, the gating network $G$ produces a distribution on the $n$ experts based on the input and assigns a weight $G(x)_{i}$ to the output of the $i^{th}$ expert. The weight can also be interpreted as the prior probability that the expert $i$ could generate the desired prediction. The final output of the model is a weighted sum of the outputs of all experts.

Subsequently, the spatial and temporal gating network will be introduced.
Comparing with MoE, the ST-ExpertNet also consists of a set of $K$ "expert networks", $E_{1}, \cdots, E_{K}$, and two "gating networks" $G_s, G_t$. $G_s$ is to control the spatial structure output and is formally named as \textit{Spatial Gating Networks}.  on the other hand, $G_t$ is to adjust the expert temporal signal and formally named as \textit{Temporal Gating Networks}.  \figref{fig:model} shows an overview of the ST-ExpertNet architecture. The experts are themselves neural networks, each with their own parameters. 



\noindent\textbf{Spatial Gating Network.}
Different from \equref{equ:output} which generates and sets $G_s$ as the weight parameter of experts, however, the objective of the spatial gating networks $G_s$ is to adjust the prediction of experts in spatial aspect. It can automatically disentangle mixed crowd flows into basic patterns and identify the current dominant pattern of each region. Then, different regions will be assigned to different specialized experts through the attention mechanism according to their dominant flow patterns. In our design, we try to avoid the ambiguous scenario in which expert $i$ outputs tiny values for region $a$, which means that it does not charge this region, but receives high attention values from $G_s$. To comprehensively handle the output with $G_s$ and experts, this paper deploys the attention model by combining $G_s$ and $E$ here to ensure the consistency of the gating and expert network, which can be written as
\begin{align}
	a_i &= \frac{{\rm exp}(G_s(X_t)_{i} \odot E_{i}(X_t))}{{\rm exp}(\sum_{i=1}^{n} G_s(X_t)_{i}\odot E_{i}(X_t))}, \quad i= 1 \cdots n, \label{equ:attention}
\end{align}
where ${\rm \textit{exp}}$ denotes the exponential operation. Therefore, by adopting attention layer, the spatial gating network final output can be written as 

\begin{align}
	e_i = a_i E_{i}(X_t),
	\label{equ:expert&attention}
\end{align}
and $e_i$ presents the final output of expert $E_{i}$


\noindent\textbf{Temporal Gating Network.}
The task of crowd flow prediction is a problem of both spatial and temporal. Time is also an important factor that affects the flow of different patterns. For instance, on weekdays, the morning outflow of the working pattern in a residential area is significantly larger than on weekends. However, the expert network, due to its self-characteristics, cannot always perceive this change and is likely to also produce high values for the morning outflow of working pattern on weekends. Thus, we consider it to be essential adding a module to help expert networks capture and make a smooth transition over these temporal changes.
Inspired by PixelCNN \cite{oord2016conditional} which uses a gated activation unit to control the final time series signal, we here design the temporal gating network $G_t$ to modify the amplitude of time series output of $y_i$.  To sum up, the final output of  ST-ExpertNet can be written as follows:
\begin{align}
	\hat{Y}_{t+1}=\text{Tanh}(\sum_{i=1}^{K} e_i) \odot \sigma(G_t(X_t)),\label{equ:total_output}
\end{align}
where $\odot$ denotes an element-wise multiplication operator. The $Tanh$ localizes the input between the range of (-1, 1) and $\sigma$ denotes the Sigmoid function to limit $G_t(x)$' s value into (0, 1). Note that \equref{equ:total_output} is only the general form of ST-ExpertNet's output. For multichannel cases, for example, input with inflow and outflow channels, \equref{equ:total_output} will act on them independently. 

\subsection{Experts' Regularization }\label{regularization}
As aforementioned criteria of \textit{sparsity} and \textit{preciseness}, here we introduce two regularization strategies: \textit{Expert Responsibility Loss} and \textit{Expert Inter Discrepancy Loss}. The former intends to force the experts taking responsibility for their own prediction, and the latter aims to drive the experts learning different flow patterns.

\noindent\textbf{Expert Responsibility Loss}
	An important issue of MoE is that we need to ensure the localization of each expert in their own subtasks. We try to avoid the scenario that $G_s$ assigns the entertaining area which is now hold by entertaining pattern to expert $i$, but expert $i$ refuses to focus on it and try to produce high value in the office area. This will lead to a great inconsistency between the expert's output and the attention mechanism. 
	To solve this problem, a punitive strategy named \textit{experts responsibility loss} has been proposed to restrict and control the high confidence between $G_s(x)_{i}$ and $E_{i}(x)$, and experts should assume responsibility for their own behavior. We set up \textit{expert responsibility loss} as $L_{er}$ intending to minimize the gap between expert output and their responsible ground-true flow. Suppose we have $K$ experts and their output $E(x)$, the general version of the error function is 
	\begin{align}
		L_{er}=\sum_{i=1}^{K} a_{i}\left(Y_{t+1}- H_i\right)^{2},\label{equ:loss_gen}
	\end{align}
	where $Y_{t+1}$ denotes the ground truth flow tensor, and $H_i=\sigma(G_t(x)) \odot \text{Tanh}(E_{i}(X_t))$. In this error function, the weights of each expert are updated based on their own error to the prediction target and thus, decouples from the influence of other experts. If we use a gradient descent method to train the network with \equref{equ:loss_gen}, the network will tend to dedicate a single expert to each basic flow pattern. \equref{equ:loss_gen} works in practice, however, according to \cite{jacobs1991adaptive}, if we further introduce negative log probability and mixture of Gaussian model into the function, it might give better performance. 
	The new formulation of $L_{er}$ is defined as:
	\begin{small}
		\begin{align}
			L_{er}= -\log \sum_{i=1}^{K} a_{i} \exp(-\frac{||Y_{t+1}-H_i ||^{2}}{2} ).\label{equ:loss_er}
		\end{align}
\end{small}

To evaluate these two functions, we start by analyzing their deviations with respect to the $i$-th expert. From \equref{equ:loss_gen}, we get
	\begin{small}
		\begin{align}
			\frac{\partial L_{er}}{\partial E_{i}(X_t)}=-2 a_iH_i^{'}\left(Y_{t+1}- H_i\right),\label{equ:d1}
		\end{align}
	\end{small}
	where $H_i^{'}=\sigma(G_t(x))\odot (1-\text{Tanh}^2(E_{i}(X_t)))$, while from \equref{equ:loss_er}, we deduce
	\begin{footnotesize}
		\begin{align}
			\frac{\partial L_{er}}{\partial E_{i}(X_t)}=-H_i^{'}\left[\frac{a_{i} e^{-\frac{1}{2}\left\|Y_{t+1}- H_i\right\|^{2}}}{\sum_{j} a_{j} e^{-\frac{1}{2}\left\|Y_{t+1}- H_i\right\|^{2}}}\right]\left(Y_{t+1}- H_i\right).\label{equ:d2}
		\end{align}
	\end{footnotesize}
	It is obvious from the deviations that both functions update an expert's weights based on their individual error. However, in \equref{equ:d1}, we use  term $a_i$ to be the weight update factor of each expert, while in \equref{equ:d2}, the weighting term further considers the ratio of an expert's error value to the total error. This unique feature allows network trained with the new error function to find the most suitable expert for a specific subtask more quickly, especially in early training stages \cite{masoudnia2014mixture}. Therefore, in this paper, we choose \equref{equ:loss_er} to be the Expert Responsibility Loss.



\begin{figure}[t]
	\centering
	\includegraphics[width=1\linewidth]{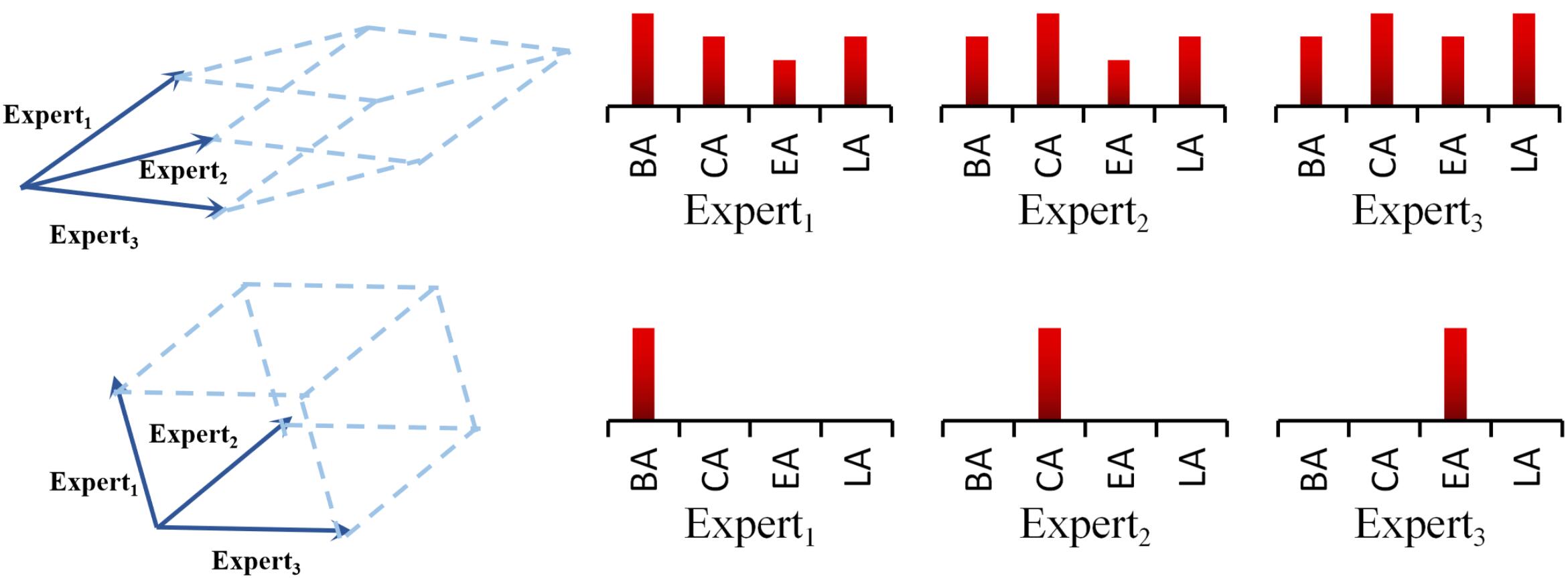}
	\caption{Here shows an example of not punishing the expert discrepancy (upper) and its counterpart (lower). The histogram indicates to what extent an expert charges a flow pattern. $BA$, $CA$, $EA$ and $LA$ refer to the pattern of Business Area, Commuting Area, Entertainment Area and Living Area respectively.}
	\label{fig:ED}
\end{figure}

\noindent\textbf{Experts Inter Discrepancy Loss.} Dividing the original task into several subtasks and localizing each expert into their corresponding subspaces are the key idea of MoE. Previous work has also shown that promoting the diversity of experts can improve the performance of the network\cite{krogh1995validation}. Thus, we intend to punish the experts inter discrepancy and try forcing the experts to learn from non-overlapping problem spaces. Inspired by the mathematical theory of the determinant point process\cite{kulesza2012determinantal} and the design of the ADP regularizer \cite{pang2019improving}, we proposed \textit{experts inter discrepancy loss} to increase the gap between experts. The loss can be defined as
\begin{align}
	L_{eid}= - det(V^TV).
\end{align}

According to the variable definition of \equref{equ:attention} and \equref{equ:expert&attention}, we can unfold $V$ as
\begin{align}
	V=[\bar{g_1}\widetilde{e}_1,\bar{g_2}\widetilde{e}_2, ... ,\bar{g_n}\widetilde{e}_n].
\end{align}

Here $\bar{g_i}$ represents the mean value of $i^{th}$ $G_s(x)$ and $\widetilde{e}_i$ refers to the flattened one-dimensional vector of the product of $i^{th}$ $G_s(x)$ and $i^{th}$ $E(x)$ under $L_2$ normalization. For sparsity, we only choose the experts among the top $n$ of attention values.  

It can be noticed that different from \cite{pang2019improving}'s ADP regularizer where all the columns of $V$ should be normalized vectors, we make an improvement by including the mean value of the corresponding attention to make each column vector has its own ``length''. This design further ensures the sparsity of experts and is more suitable for our model because we hope that each expert can have a similar mean value of attention, which means that every expert has its own subtask instead of leaving a few experts with nothing to do.

A theory \cite{Matrixmathematics} has stated that if $A \in \mathbb{R}^{n \times m}$ is a matrix and $rank(A) = m$, we have 
\begin{align}
	volume(\delta)= [det(V^TV)]^{1/2},
\end{align}
from which $\delta$ denotes the geometrical body spanned by $V$ 's column vector set. Therefore, we can have a geometrical illustration of the expert inter discrepancy loss at \figref{fig:ED}. The volume spanned by the vector set indicates the divergence between experts. We intend to maximize the volume to force each expert to pay attention to different flow patterns of the city. The histogram in \figref{fig:ED} shows that before we punish the expert discrepancy, different experts might make nearly equal contributions to the same flow pattern. However, after applying the punishment, each expert now has its own pattern to charge, and this agrees with our initial motivation.

\subsection{Training Steps}\label{training_step}
To sum up the previous discussion in Sections \ref{sec:Formulation}  and \ref{sec:Methodology},  we solve the problem in \equref{equ:prediction} by proposing the ST-ExpertNet with $K$ experts and 2 gating networks from spatial and temporal aspects, solve the problem in \equref{equ:lambda1} by proposing \textit{experts responsibility loss}, and solve the problem in \equref{equ:lambda2} by proposing \textit{experts inter discrepancy loss}. We need to train $K$ expert models and two gating networks, which take input from multisource information and concatenate them as $X_t$.  Our ST-ExpertNet can be trained to predict $y_{t+1}$ by minimizing the mean square error between the predicted flow matrix and the true flow matrix. The total loss function is written as 
\begin{small}
	\begin{align}
		\mathcal{L}(\theta)=(1-\lambda_{er}-\lambda_{eid})\left\|\hat{Y}_{t+1}-Y_{t+1}\right\|_{2}^{2}+ \lambda_{er}L_{er}+\lambda_{eid}L_{eid},
	\end{align}  
\end{small}

where $\theta$ are all the learnable parameters in ST-ExpertNet and $\lambda_{er}>0$ and $\lambda_{eid}>0$ denote the intensity of punishment for $L_{er}$ and $L_{eid}$ respectively.

To examine the effect of \textit{experts inter discrepancy loss}, we conduct some experiments under different settings of $\lambda_{eid}$ while keeping $\lambda_{er}$ fixed as $10^{-2}$. 

%
\section{Experiment}\label{sec:exps}
In this section, we will conduct widely experiments on ST-ExpertNet with four real world public crowd flow datasets: TaxiBJ, TaxiNYC, BikeNYC-I, as well as BikeNYC-II. 
The quantitative and qualitative experiments will be brought into effect to demonstrate the superiority of our proposed method. 

\subsection{Datasets}

\begin{table*}[t]
	\centering
	\footnotesize
	\caption{ Effectiveness Evaluation on TaxiBJ and TaxiNYC }\label{tab:result1}
	\begin{tabular}{|l|c|c|c|c|c|c|c|c|c|c|}
		\hline & \multicolumn{5}{|c|} { \textbf{TaxiBJ} } & \multicolumn{5}{|c|} { \textbf{TaxiNYC} } \\
		\hline Model & MSE & RMSE & MAE  & MAPE & Improved & MSE & RMSE & MAE & MAPE & Improved \\
		\hline HistoricalAverage & $2025.328$ & $45.004$ & $24.475$ & $8.04 \%$ & $\backslash$ &$463.763$ & $21.535$ & $7.121$ & $4.56 \%$ &$\backslash$\\
		CopyYesterday & $1998.375$ & $44.703$ & $22.454$ & $8.14 \%$ & $\backslash$ & $1286.035$ & $35.861$ & $10.164$ & $5.78 \%$ &$\backslash$\\
		CNN & $554.615$ & $23.550$ & $13.797$ & $8.46 \%$ &$\backslash$ & $280.262$ & $16.741$ & $6.884$ & $8.08 \%$&$\backslash$ \\
		ConvLSTM & $370.448$ & $19.247$ & $10.816$ & $5.61 \%$ &$\backslash$ &$147.447$ & $12.143$ & $4.811$ & $5.16 \%$&$\backslash$ \\
		ST-ResNet & $349.754$ & $18.702$ & $10.493$ & $5.19 \%$ & $\backslash$&$133.479$ & $11.553$ & $4.535$ & $4.32 \%$ &$\backslash$\\
		DMVST-Net & $415.739$ & $20.389$ & $11.832$ & $5.99 \%$ & $\backslash$&$185.601$ & $13.605$ & $4.928$ & $4.49 \%$ &$\backslash$\\
		PCRN & $355.511$ & $18.855$ & $10.926$ & $6.24 \%$ & $\backslash$&$181.091$ & $13.457$ & $5.453$ & $6.27 \%$&$\backslash$ \\
		CNN-ExpertNet & 342.737& 18.513& 10.356&  5.28\%&\textbf{62\%}& 114.717  &  10.711  &   4.186&   4.50\% &\textbf{144\%}  \\
		ConvLSTM-ExpertNet & \textbf{328.250}& \textbf{18.117}& \textbf{10.079}& \textbf{4.95}\% & 12\% &\textbf{109.445}& \textbf{10.503}&  \textbf{4.104}& \textbf{3.93}\%& 36\% \\
		ST-ResNet-ExpertNet & {330.258}& {18.173}&  {10.283}&   {5.05}\% &6\% &111.445&  10.557&   4.264&   4.04\% &20\% \\
		\hline
	\end{tabular}
	
\end{table*}

\begin{table*}[t]
	\centering
	\footnotesize
	\caption{ Effectiveness Evaluation on BikeNYC-I and BikeNYC-II }\label{tab:result2}
	\begin{tabular}{|l|c|c|c|c|c|c|c|c|c|c|}
		\hline & \multicolumn{5}{|c|} { \textbf{BikeNYC-I} } & \multicolumn{5}{|c|} { \textbf{BikeNYC-II} } \\
		\hline Model & MSE & RMSE & MAE  & MAPE & Improved & MSE & RMSE & MAE & MAPE & Improved \\
		\hline HistoricalAverage & ~$245.743$~ & $15.676$ & $4.882$ & $5.45 \%$ & $\backslash$ &~$23.757$~ & ~$4.874$~ & $1.500$ & $3.30 \%$ &$\backslash$\\
		CopyYesterday & ~$241.681$~ & $15.546$ & $4.609$ & $5.36 \%$ & $\backslash$ & ~$54.054$~ & ~$7.352$~ & $1.995$ & $3.94 \%$ &$\backslash$\\
		CNN & ~$145.549$~ & $12.064$ & $4.088$ & $5.82 \%$ &$\backslash$ & ~$20.351$~ & ~$4.511$~ & $1.574$ & $3.98 \%$&$\backslash$ \\
		ConvLSTM & ~$43.765$~ & $6.616$ & $2.412$ & $3.90 \%$ &$\backslash$ &~$10.076$~ & ~$3.174$~ & $1.133$ & $2.90 \%$&$\backslash$ \\
		ST-ResNet & ~$37.279$~ & $6.106$ & $2.360$ & $3.72 \%$ & $\backslash$&~$10.182$~ & ~$3.191$~ & $1.169$ & $2.86 \%$ &$\backslash$\\
		DMVST-Net & ~$63.849$~ & $7.990$ & $2.833$ & $3.93 \%$ & $\backslash$&~$12.397$~ & ~$3.521$~ & $1.287$ & $2.97 \%$ &$\backslash$\\
		PCRN & ~$130.878$~ & $11.440$ & $3.790$ & $4.90 \%$ & $\backslash$&~$10.893$~ &~ $3.300$ ~& $1.185$ & $3.04 \%$&$\backslash$ \\
		CNN-ExpertNet & ~36.068~ & 6.006& 2.244&  3.60\%& \textbf{304\%}& ~ 8.577~  &  ~2.929~  &   1.100 &   2.42\% &\textbf{137\%}  \\
		ConvLSTM-ExpertNet & ~\textbf{35.338}~&  \textbf{5.945}&  \textbf{2.153}&   \textbf{3.48}\% & 24\% &~\textbf{8.574} ~&~ \textbf{2.928}~&  \textbf{1.061}& \textbf{2.40}\%& 18\% \\
		ST-ResNet-ExpertNet & ~36.722~& 6.073&  2.376&   3.67\% &1.52\% &~8.806~&  ~2.968~&   1.174&   2.88\% &16\% \\
		\hline
	\end{tabular}
\end{table*}

\noindent\textbf{TaxiBJ}: Trajectory GPS data is collected in Beijing \cite{zhang2017deep} taxicab from four-time intervals: 01/07/2013 - 30/10/2013,  01/03/2014 - 30/06/2014, 01/03/2015 - 30/06/2015, and 01/11/2015 - 10/04/2016. The map of Beijing is split into $32 \times 32$ uniform regions, and metadata features are also stored in this dataset with temperature, wind speed, weather conditions, and holidays information. 

\noindent\textbf {TaxiNYC}. TaxiNYC is the taxi in-out flow data created \cite{yao2019revisiting} from 01/01/2015 to 03/01/2015 in NYC-Taxi.
We use the holidays and time information as external factors.

\noindent\textbf {BikeNYC-I, BikeNYC-II} These datasets are published   
\cite{yao2019revisiting} and \cite{lin2019deepstn}  from  time intervals: 07/01/2016 - 08/29/2016 and  4/1/2014-9/30/2014 respectively in NYC-Bike.  I and II are used to distinguish the datasets released from different sources.

\subsection{Evaluation Metric}
We use MSE (Mean Squared Error), RMSE (Root Mean Square Error),
MAE (Mean Absolute Error) and MAPE (Mean Absolute Percentage Error) to evaluate our algorithm, which are defined as follows:
\begin{align*}
	\left\{\begin{aligned}
		MSE &=\frac{1}{N} \sum_{i=1}^{N} \left(\hat{Y}_{t+1}^{i}-Y_{t+1}^{i}\right)^{2} \\
		R M S E&= \sqrt{\frac{1}{N} \sum_{i=1}^{N}\left(\hat{Y}_{t+1}^{i}-Y_{t+1}^{i}\right)^{2}}\\
		M A E&=\frac{1}{N} \sum_{i}^{n}\left|\hat{Y}_{t+1}^{i}-Y_{t+1}^{i}\right|\\
		M A P E& =\frac{1}{N} \sum_{i}^{n}\left|\frac{\hat{Y}_{t+1}^{i}-Y_{t+1}^{i}}{\hat{Y}_{t+1}^{i}}\right|
	\end{aligned}\right.
\end{align*}
where $\hat{Y}_{t+1}^{i}$ and $Y_{t+1}^{i}$ mean the prediction value and real value of region $i$ for time interval $t+1$, and where $N$ is a total number of samples.


\subsection{Methods for Comparison }

\begin{figure*}[t]
	\centering
	\includegraphics[width=0.8\linewidth]{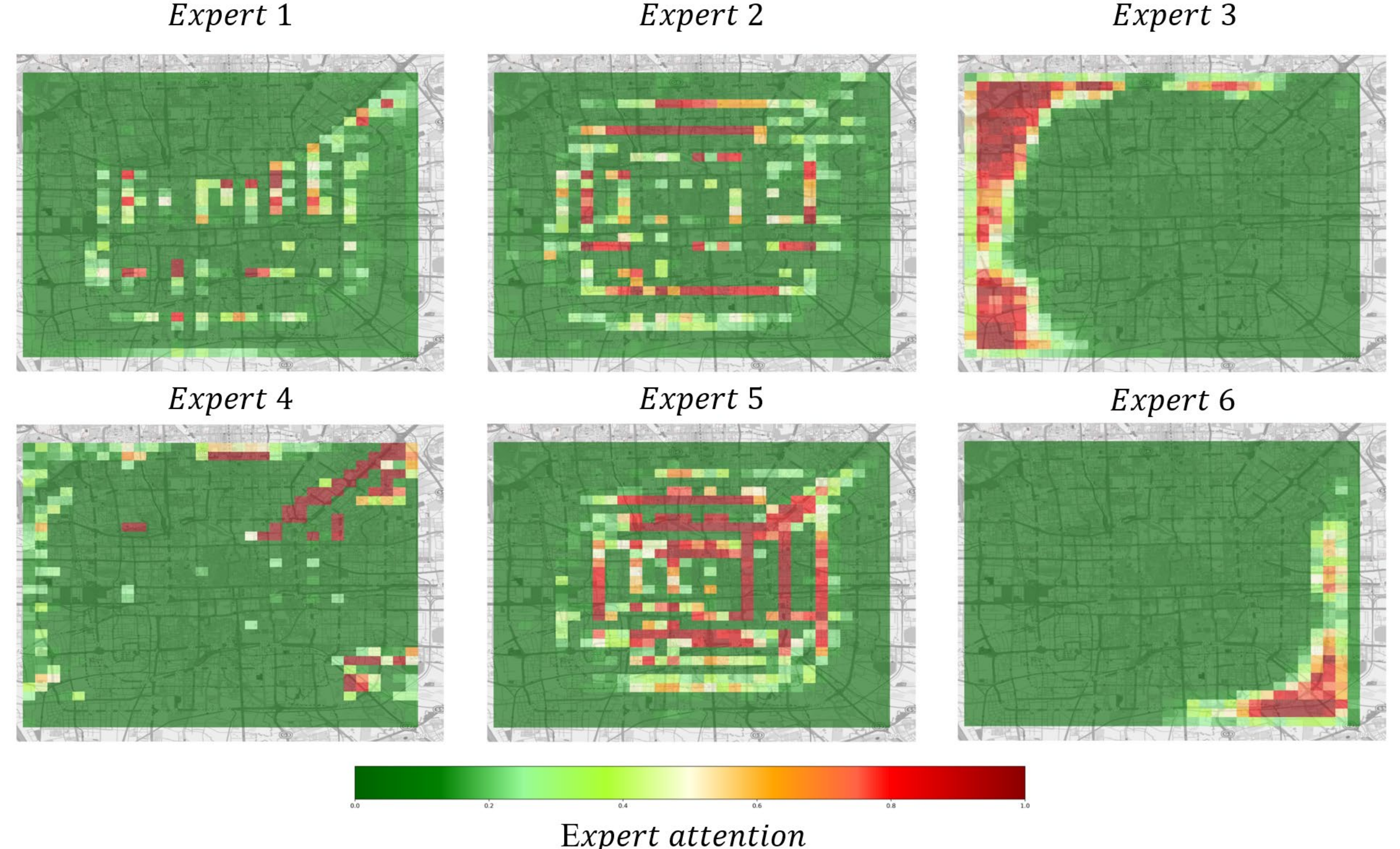}
	\caption{The ST-ExpertNet $\mathbf{G_s}$ attention visualization with spatial aspect.}
	\label{fig:attentionspatial}
\end{figure*}

We compared our model with the following methods and tuned the parameters for all the methods. Then we report the best performance in the validation set. Additionally, to test ST-ExpertNet's general applicability, we apply the architect of CNN, ConvLSTM, and ST-ResNet into our model. Here we follow the experiment setting on the benchmark work \cite{jiang2021dl} in traffic flow prediction, and the detailed implements for each ST-ExpertNet are listed as follows:
\begin{itemize}
	\item[$\bullet$] Historical average (HA): Historical average predicts the future flow by averaging the value of previous flow at the same region in the same relative time interval.
	\item[$\bullet$] Pure CNN. It is a basic popular deep learning baseline in flow prediction constructed with three CNN layers with 64 filters of the $3\times3$ kernel. The  ReLU activation function and BatchNormalization are added between two consecutive layers. 
	Pure CNNs architecture had been deployed, which only set ReLU operations in the final output. 
	
	\item[$\bullet$] ConvLSTM. The convolutional LSTM \cite{shi2015convolutional} extends the fully connected LSTM (FC-LSTM) \cite{zhao2019long} and maintains the advantage of FC-LSTM. Further, the convolutional operations staid in input-to-state and state-to-state transitions provide opportunities to capture both spatial and temporal dependency. The ConvLSTM layers maintain 32 filters of 3×3 kernel
	size. Four ConvLSTM layers are  used and the ReLU operation is set in the final output. 
	
	\item[$\bullet$] ST-ResNet \cite{zhang2017deep}: 	ST-ResNet extends the ResNet \cite{he2016deep} into the framework of traffic prediction. Moreover, the three blocks of ResNet have been designed to process the hour, day, and week patterns, respectively. The length of the sequences on closeness, period and trend are set as $l_c=3$, $l_p=1$, $l_t=1$. Each block in ST-ResNet owns 3 residual units with the 32 filters of 3×3  kernels.
	\item[$\bullet$] DMVST-Net. DMVST-Net \cite{yao2018deep} combines spatial and temporal information using local CNN and LSTM. The local CNN only capture spatial feature among nearby grids and LSTM only process the  recent time intervals.
	\item[$\bullet$] PCRN. PCRN-CRN \cite{zonoozi2018periodic} focus on capturing the 
	periodic patterns in Spatio-temporal data by using ConvGRU model \cite{ballas2015delving}. The pyramidal architecture by stacking three convolutional RNN layers  has been proposed  to learn the spatial and temporal features simultaneously.
	\item[$\bullet$] CNN-ExpertNet. We deploy the CNN into ST-ExpertNet with 10 Expert Networks. For each ExpertNet, we set up the same architecture as pure CNN.
	
	\item ConvLSTM-ExpertNet. The architect of ConvLSTM has been adopted in Expert-Net and the number of Expert Networks is set 3. Meanwhile, ExpertNet in ConvLSTM-ExpertNet also owns the same architecture as ConvLSTM.
	
	\item[$\bullet$] ST-ResNet-ExpertNet. In order to compare the performance with the original ST-ResNet, we retain the same architect of ST-ResNet and use only three expert networks. Note that the ST-ResNet-ExpertNet indicates that the three ResNet as the same as ST-ResNet  but not the three ST-ResNet blocks. For a detailed comparison with ST-ResNet and ST-ResNet-ExpertNet, please see supplementary materials.
\end{itemize}
For all different architectures of ST-ExpertNet, the $G_s$ and $G_t$ are simply designed as the same structure of pure CNN with 64 filters of $3\times3$ kernel.

\subsection{Preprocessing} 
Here, we scale all flow datasets into $[0,1]$ using the Min-Max normalization method. For MSE and RMSE, the values are re-scaled back to the normal values. Then, we use the one-hot coding method to transfer them into binary vectors for external factors of discrete value in Beijing, such as DayOfWeek, and Weekend/Weekday. For other factors of continuous values in Beijing, like temperature and wind speed, we use the Min-Max normalization method. Finally, external factors in New York are set with holidays and time information. 
	
\begin{figure}[h]
	\centering
	\includegraphics[width=1\linewidth]{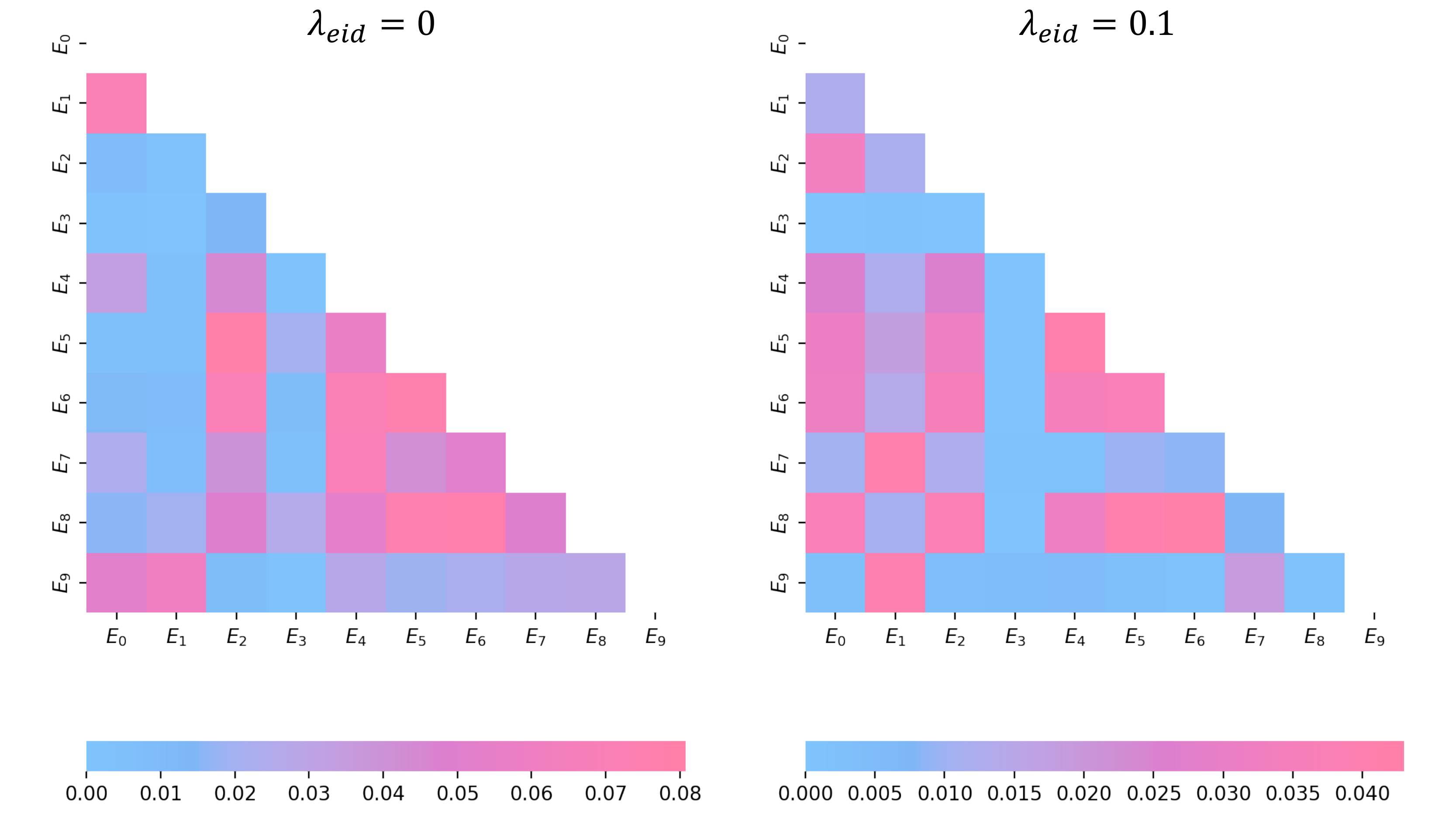}
	\caption{Quade Test analysis result at different setting of CNN-ExpertNet trained on TaxiBJ.}\label{Quade Test}
	\vspace{-0.5cm}
\end{figure}
\begin{figure}
	\centering
	\includegraphics[width=1\linewidth]{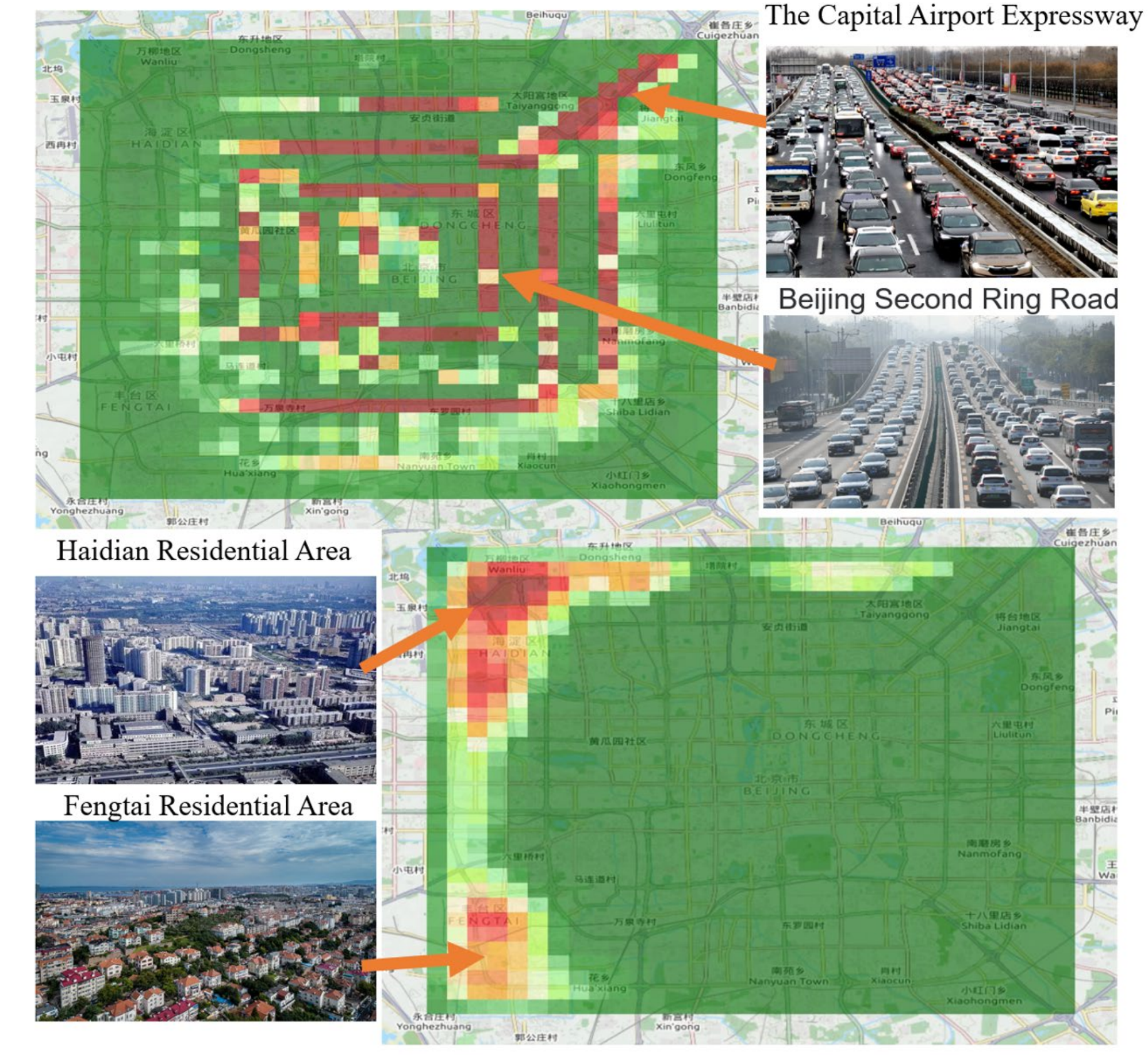}
	\caption{Visualize the attention for different functional expert in TaxiBJ.}
	\label{fig:expertvis}
\end{figure}
\subsection{Hyperparameters} 
For all methods with convolution filter, we set the kernel size with $3\times 3$. There are five additional hyperparameters: the number of experts $K$, the penalty terms $\lambda_{er}$ and $\lambda_{eid}$ for $L_{eid}$ and $L_{er}$, respectively, the length of three dependent sequences $l_{c}, l_{p}, l_{q}$ in our ST-ExpertNet. For the parameters $k ,l_{c}, l_{p}$ and $ l_{q}$, we select the best parameters in $k, l_{c},l_{p}, l_{q} \in \{1,2,3,4,5,6,7,8,9,10\}$. For the penalty term $\lambda_1$ and $\lambda_2$, we choose $\{10^{-5},10^{-4},10^{-3},10^{-2},10^{-1},1\}$. The parameter $T$ in each ST-model is set as 0.8. The uniform distribution approach in Pytorch \cite{paszke2019pytorch} is implemented in learnable parameters. We select $80 \%$ of the data as training data and $20 \%$ of the training data as a validation set, which help to   
early-stop  each model based on the best validation score. The batch size is set as 32 and Adam \cite{kingma2014adam} is used for optimization with learning rate $10^{-3}$.
\begin{figure*}[t]
	\centering
	\includegraphics[width=1\linewidth]{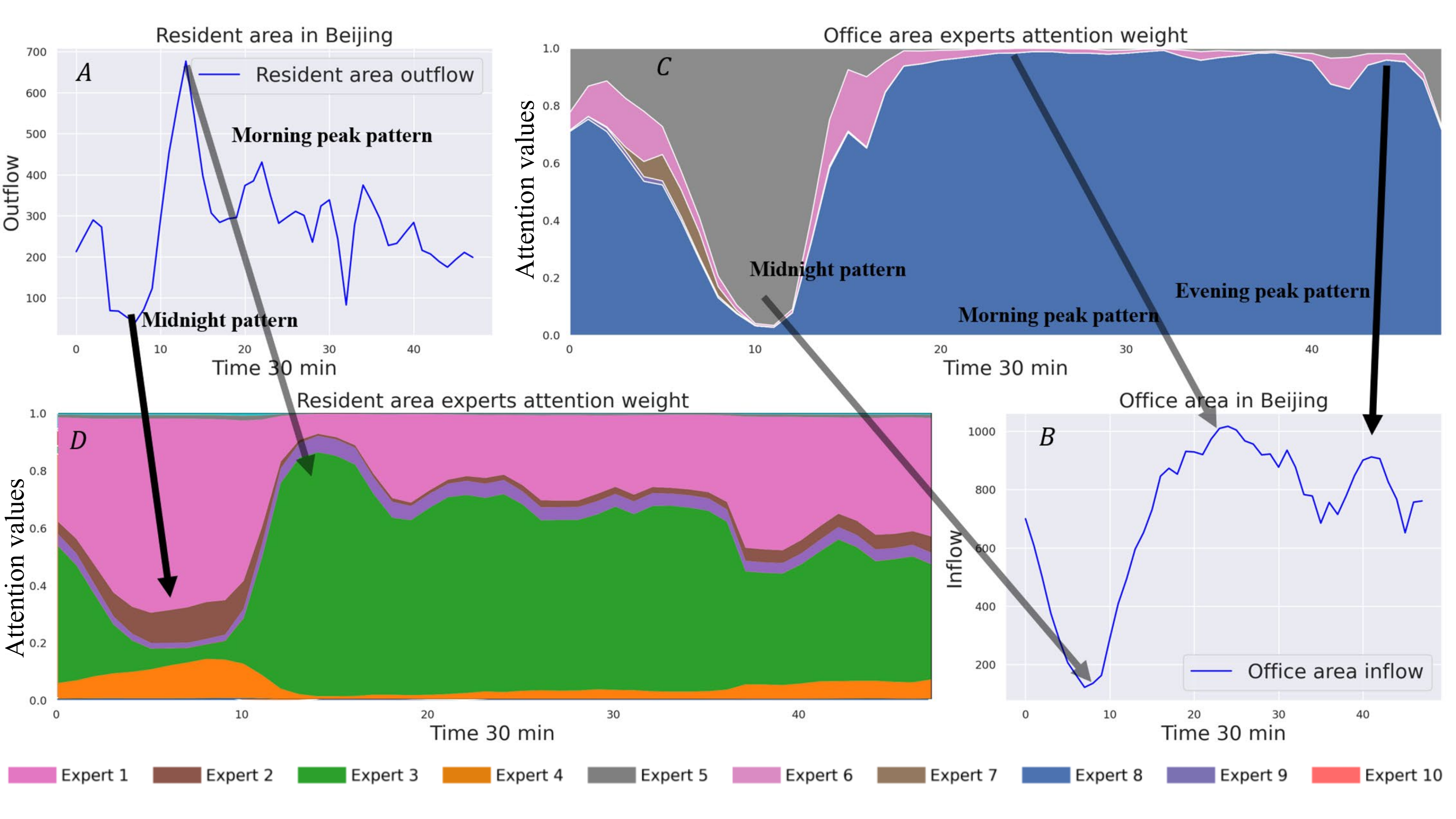}
	\caption{ \textbf{The ST-ExpertNet $\mathbf{G_s}$ attention visualization with temporal aspect.} A. \textnormal{The resident area outflow change within one day in Beijing.} B. \textnormal{The office area expert attention value.} C. \textnormal{The resident area expert attention value.} D. \textnormal{The office area inflow change within one day in Beijing.}}
	\label{fig:attentiontemporal}
\end{figure*}

\begin{figure*}[h]
	\centering
	\subfloat{\includegraphics[width=0.24\textwidth]{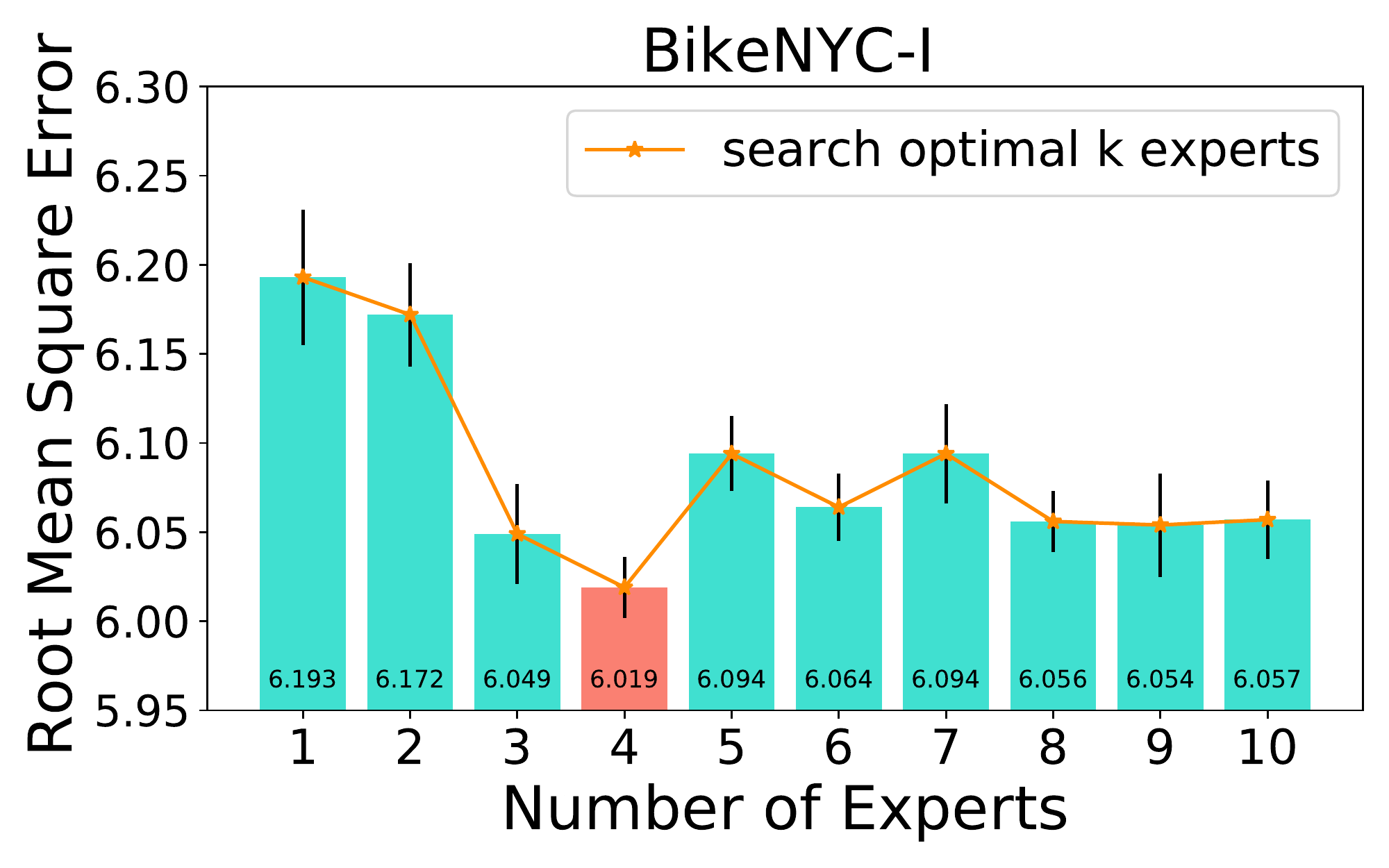}}
	\hfill
	\subfloat{\includegraphics[width=0.24\textwidth]{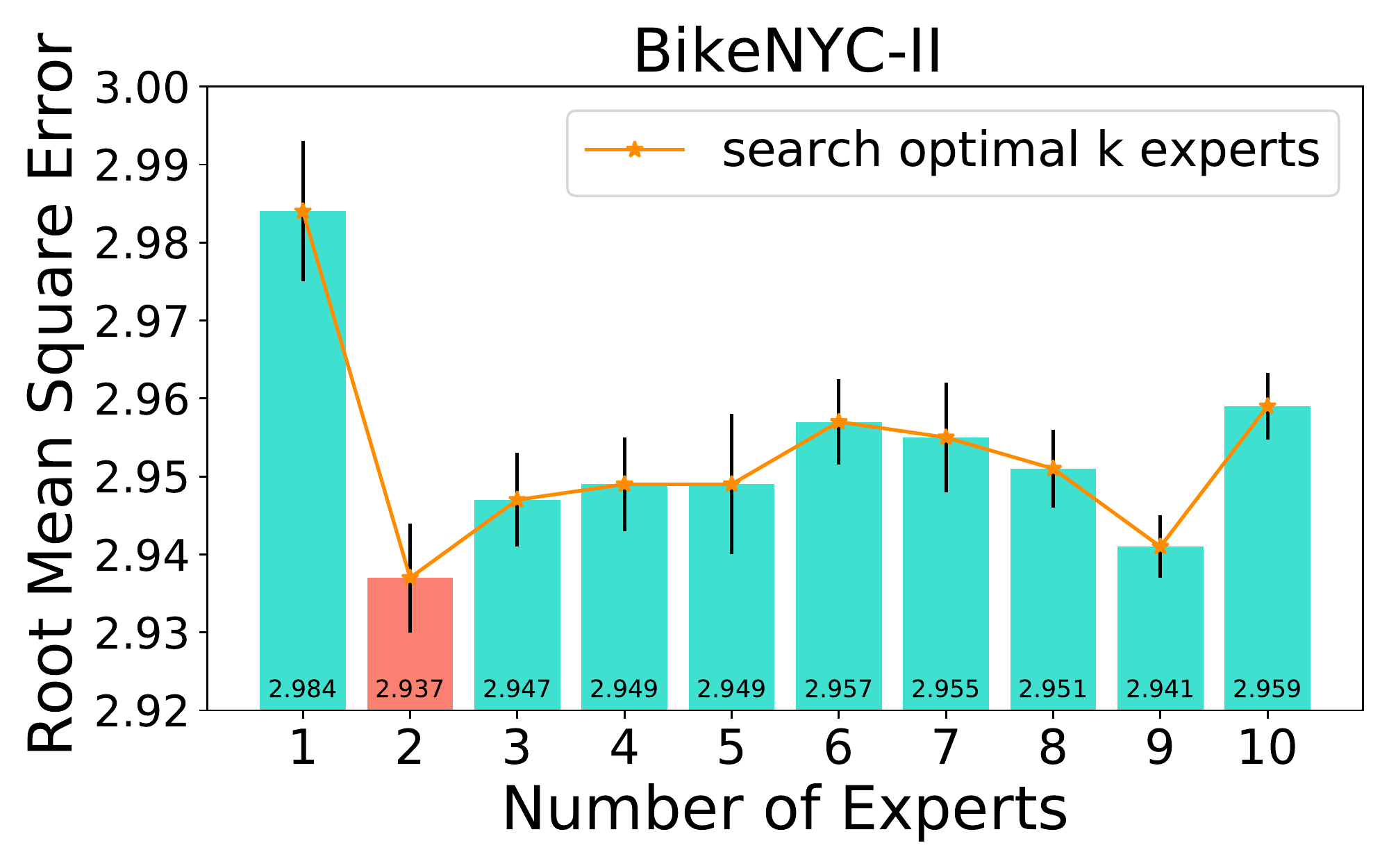}}
	\hfill
	\subfloat{\includegraphics[width=0.24\textwidth]{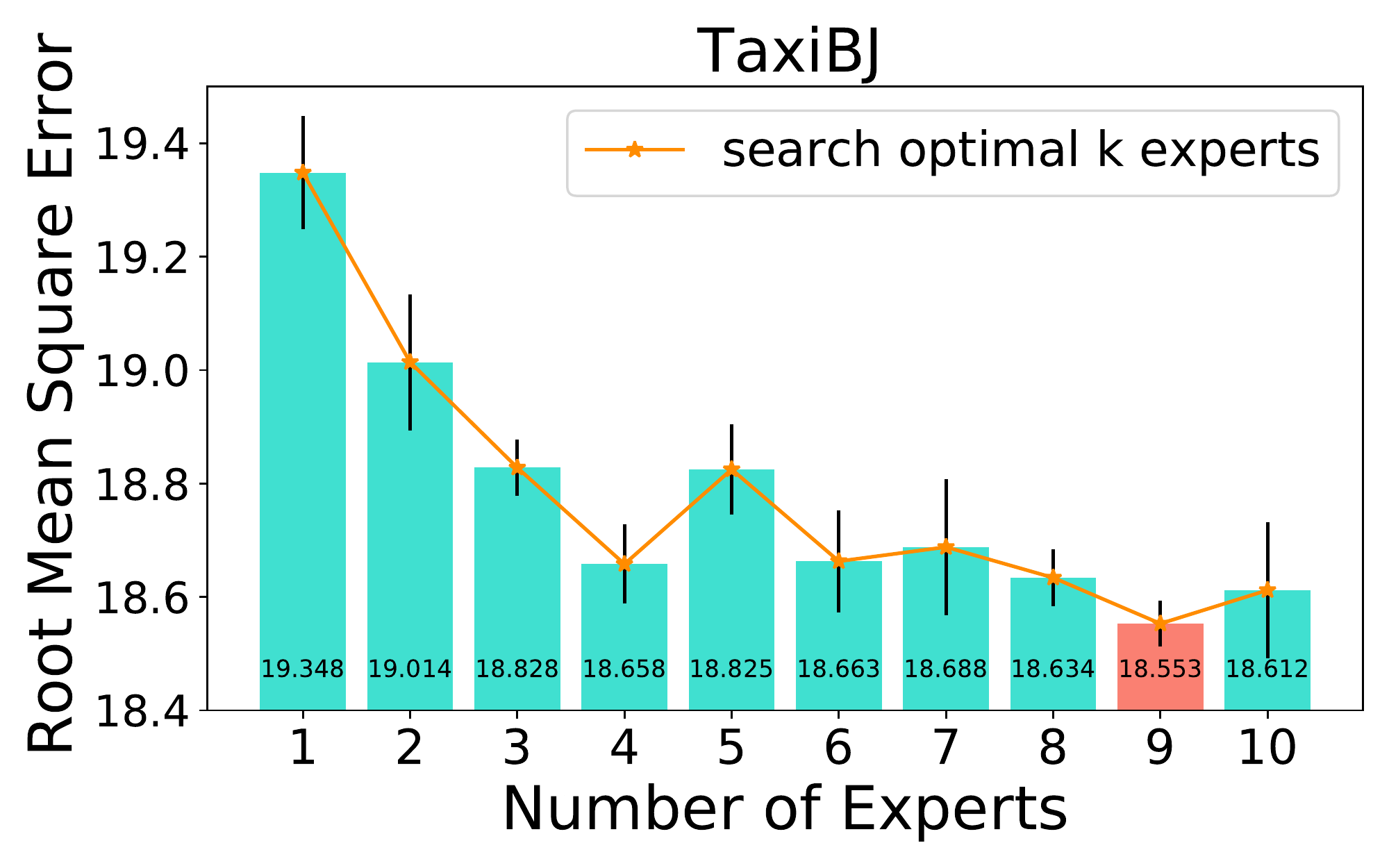}}
	\hfill
	\subfloat{\includegraphics[width=0.24\textwidth]{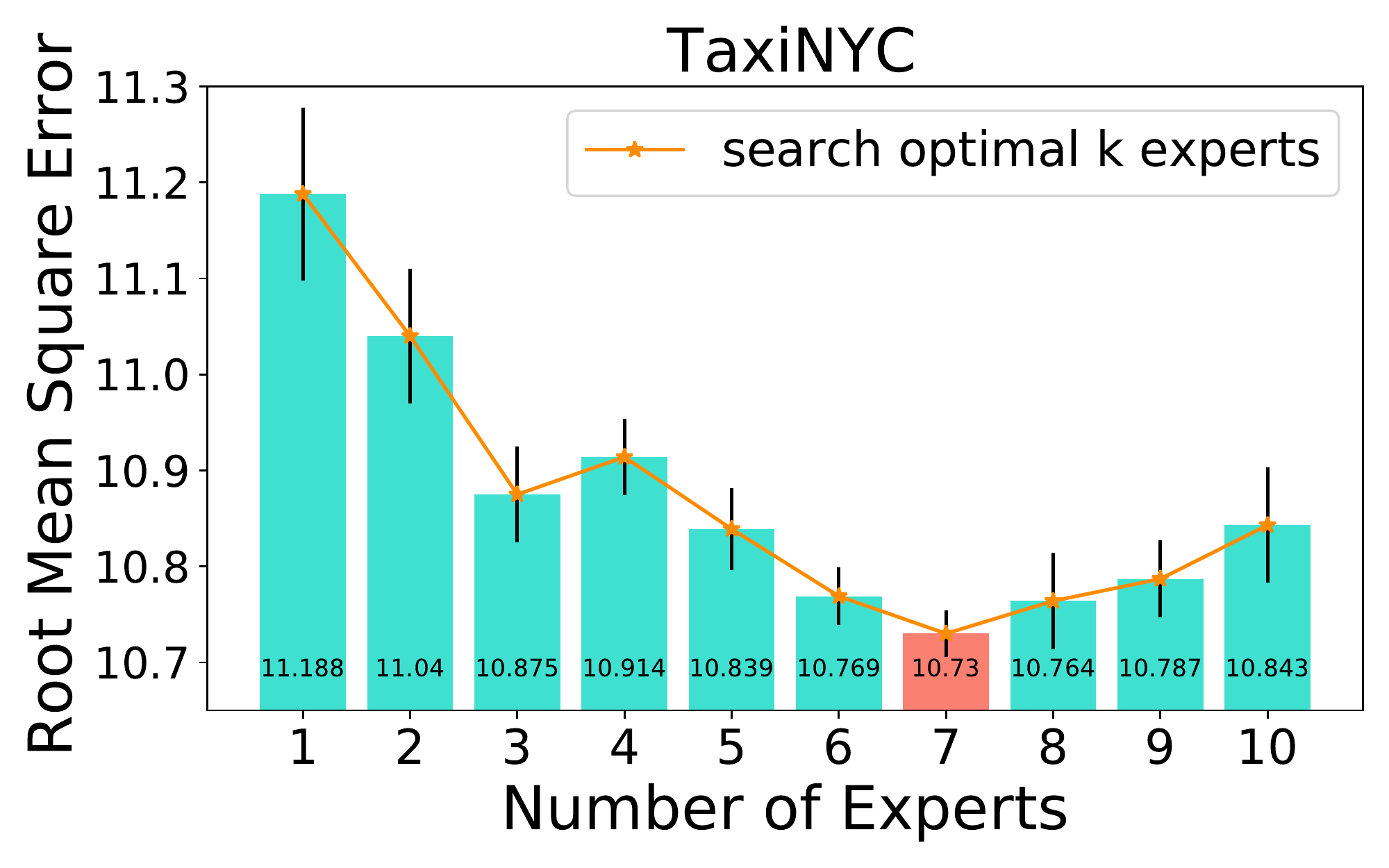}}
	
	\subfloat{\includegraphics[width=0.24\textwidth]{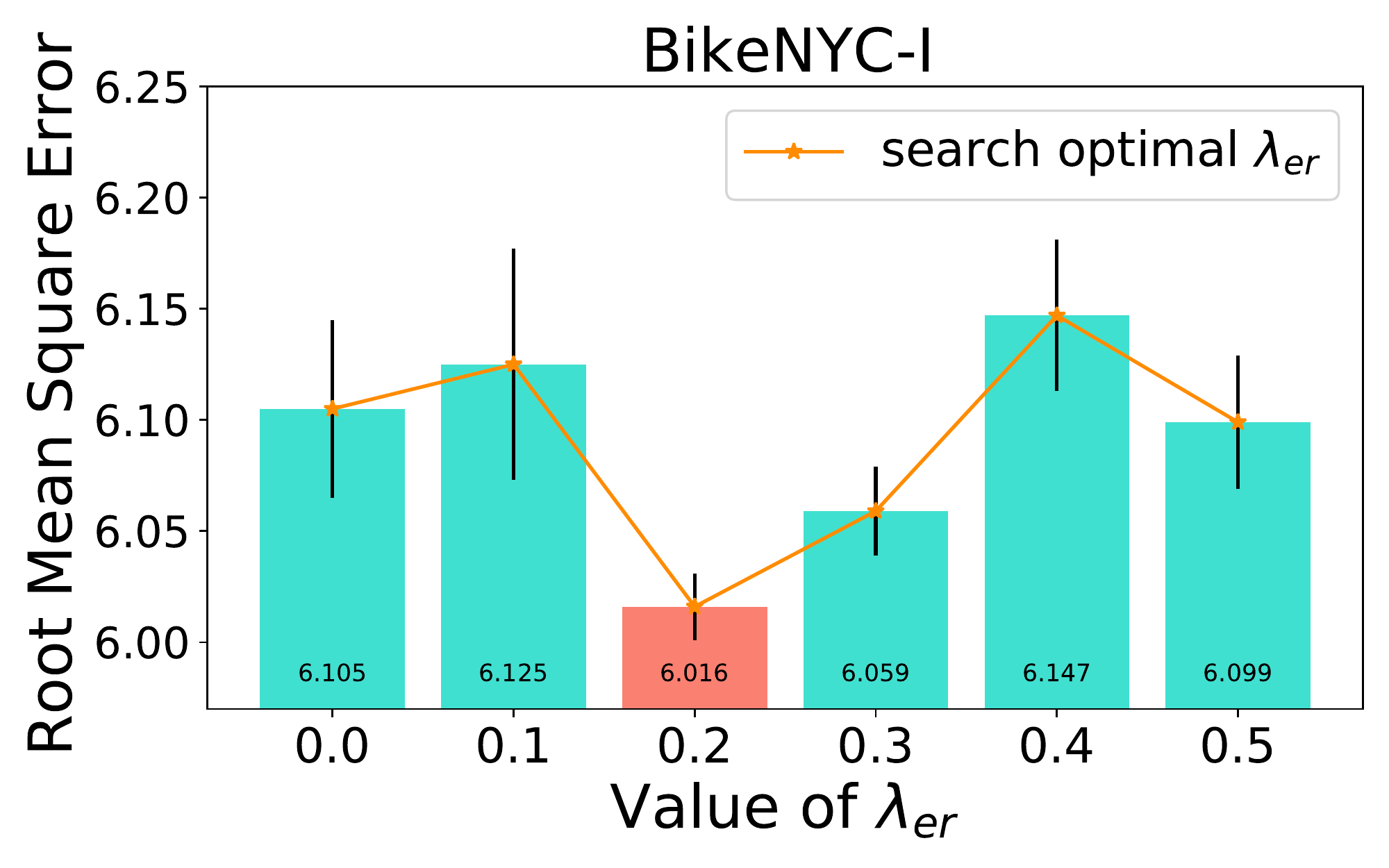}}
	\hfill
	\subfloat{\includegraphics[width=0.24\textwidth]{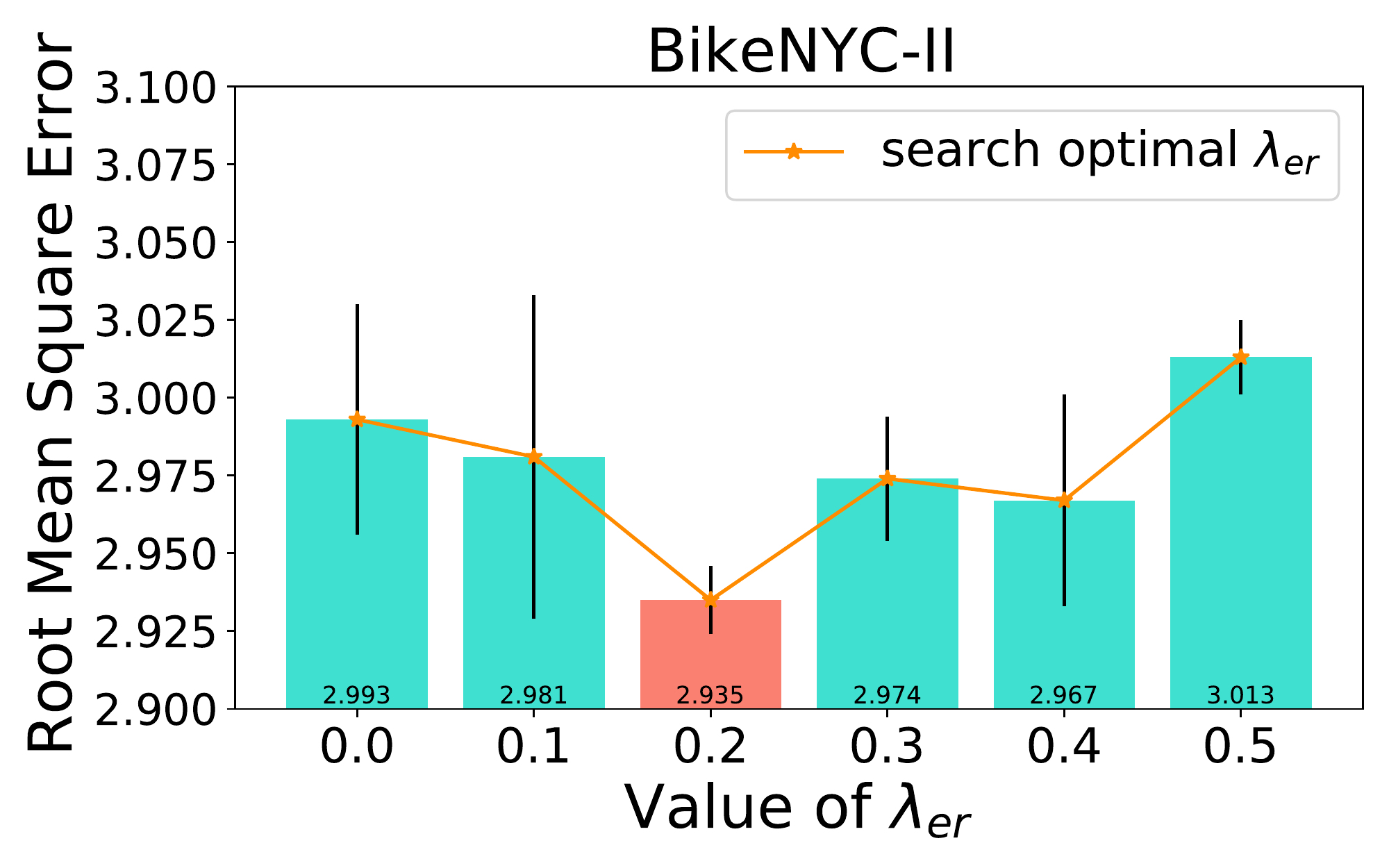}}
	\hfill
	\subfloat{\includegraphics[width=0.24\textwidth]{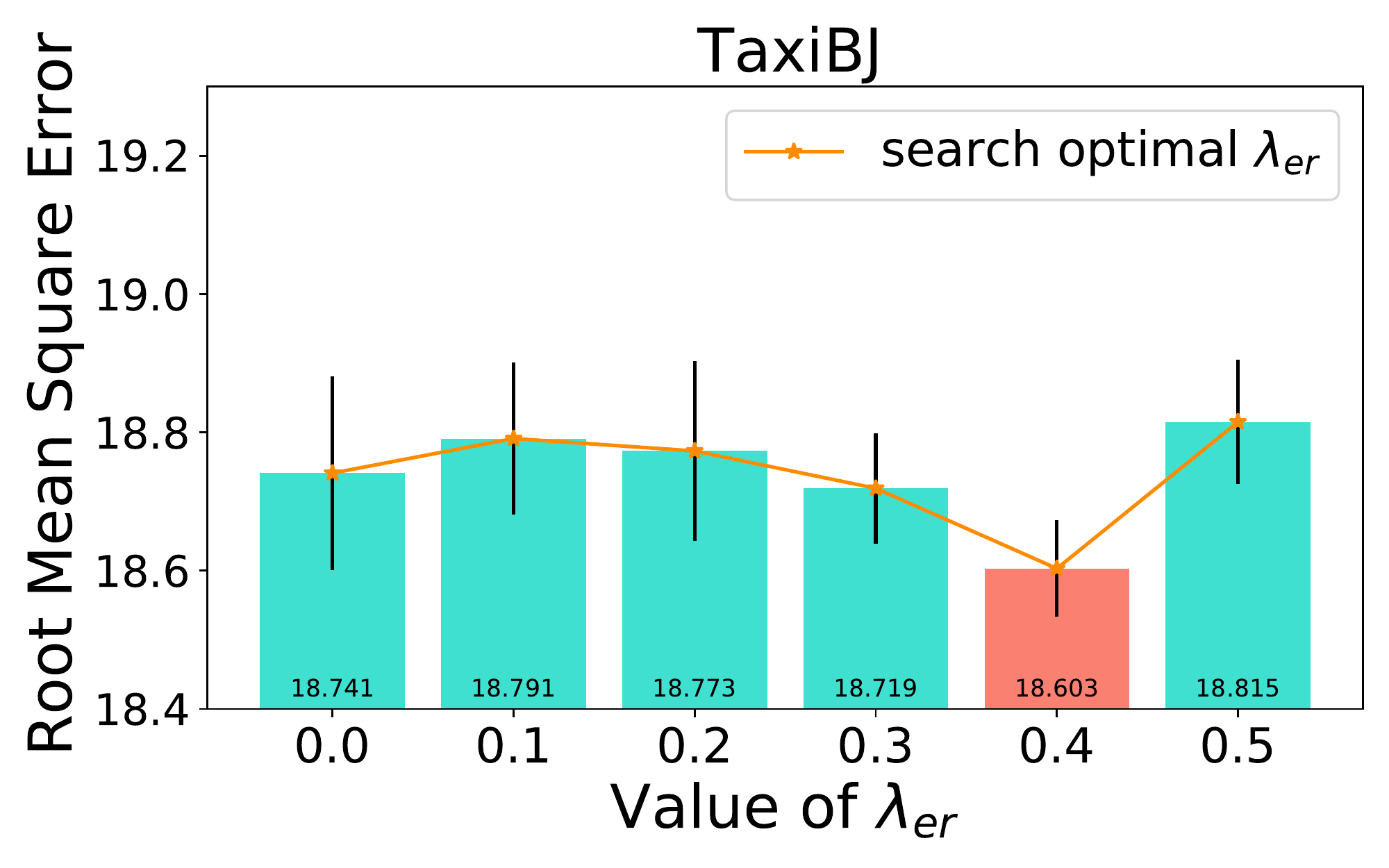}}
	\hfill
	\subfloat{\includegraphics[width=0.24\textwidth]{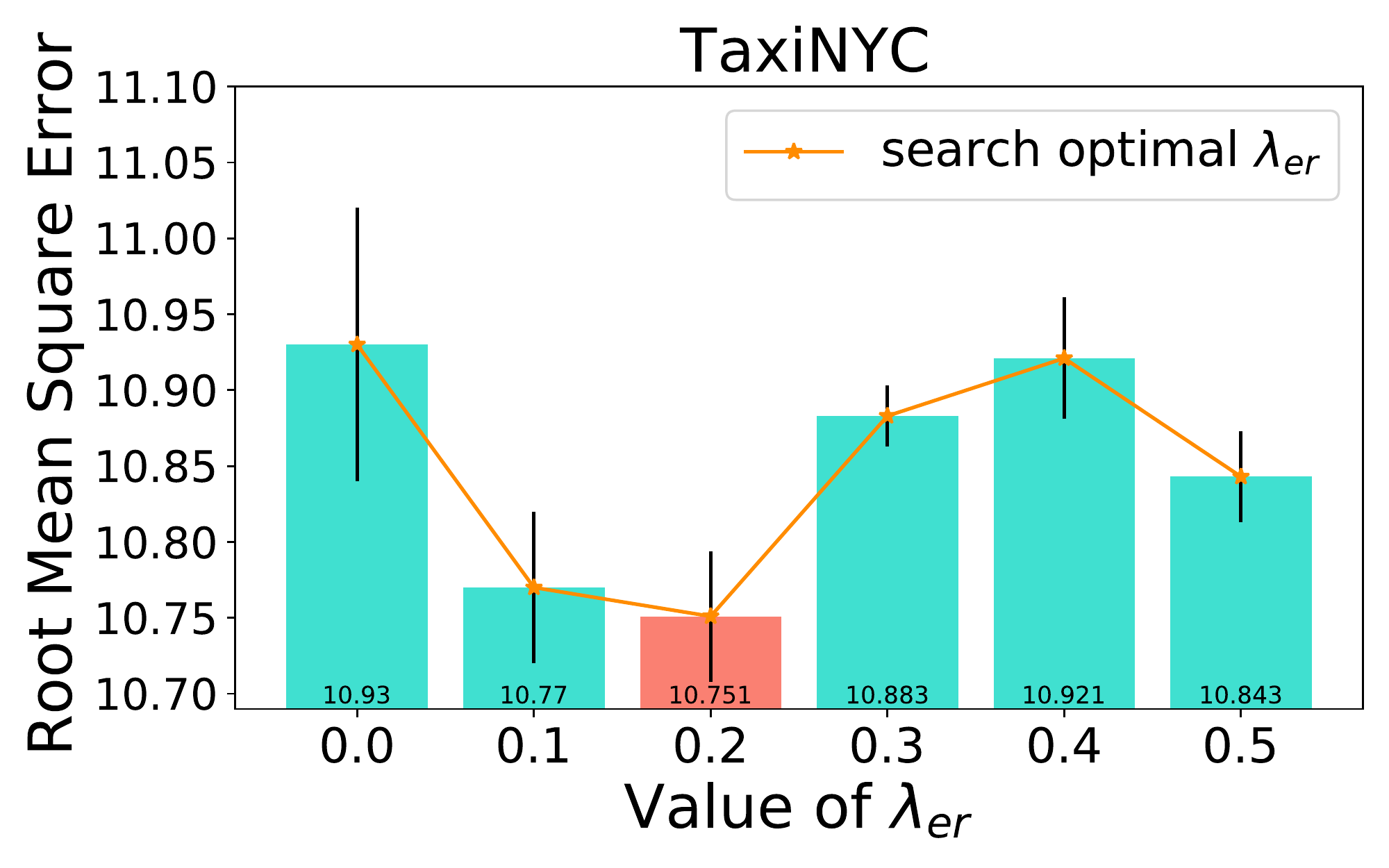}}
	
	\subfloat{\includegraphics[width=0.24\textwidth]{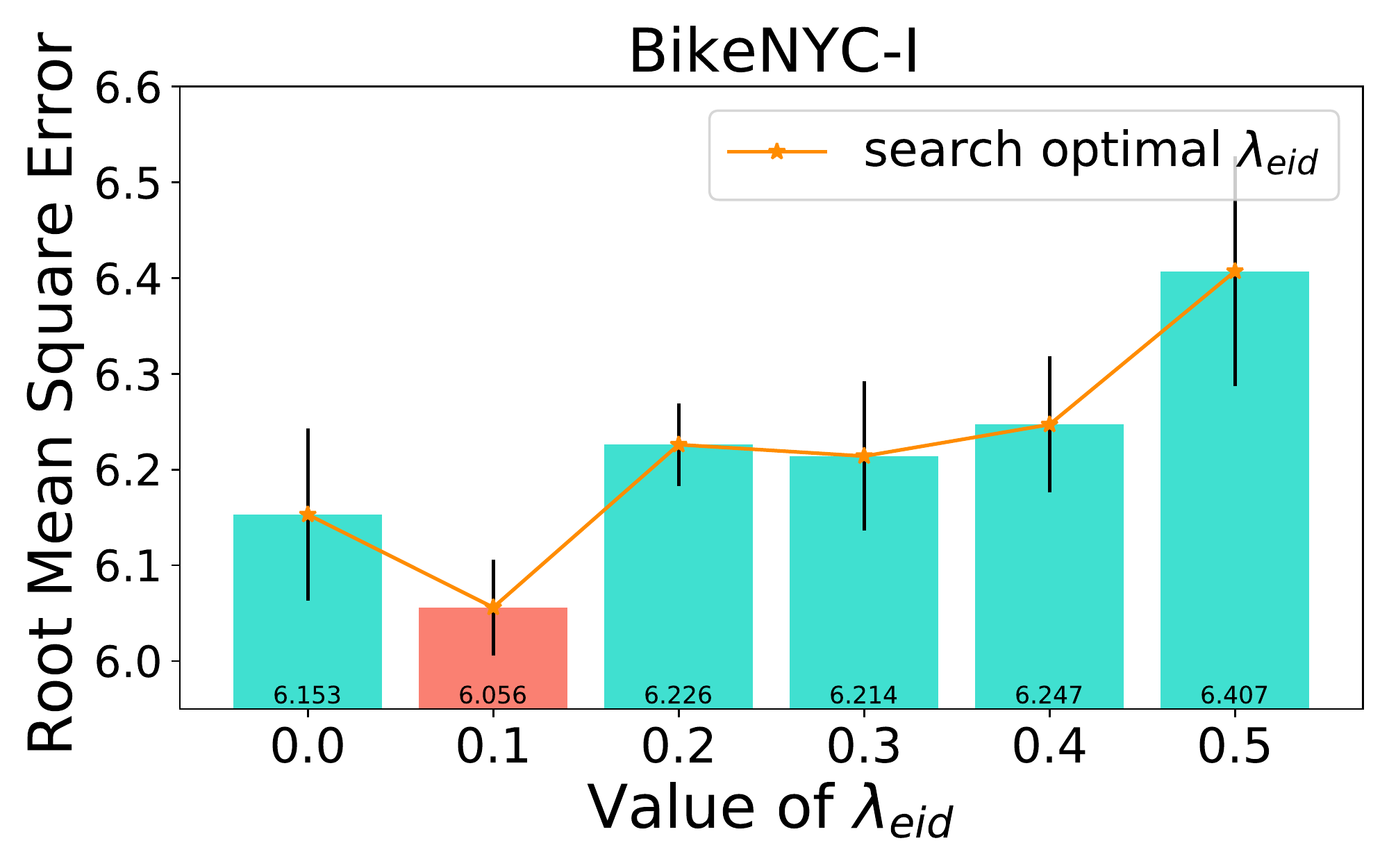}}
	\hfill
	\subfloat{\includegraphics[width=0.24\textwidth]{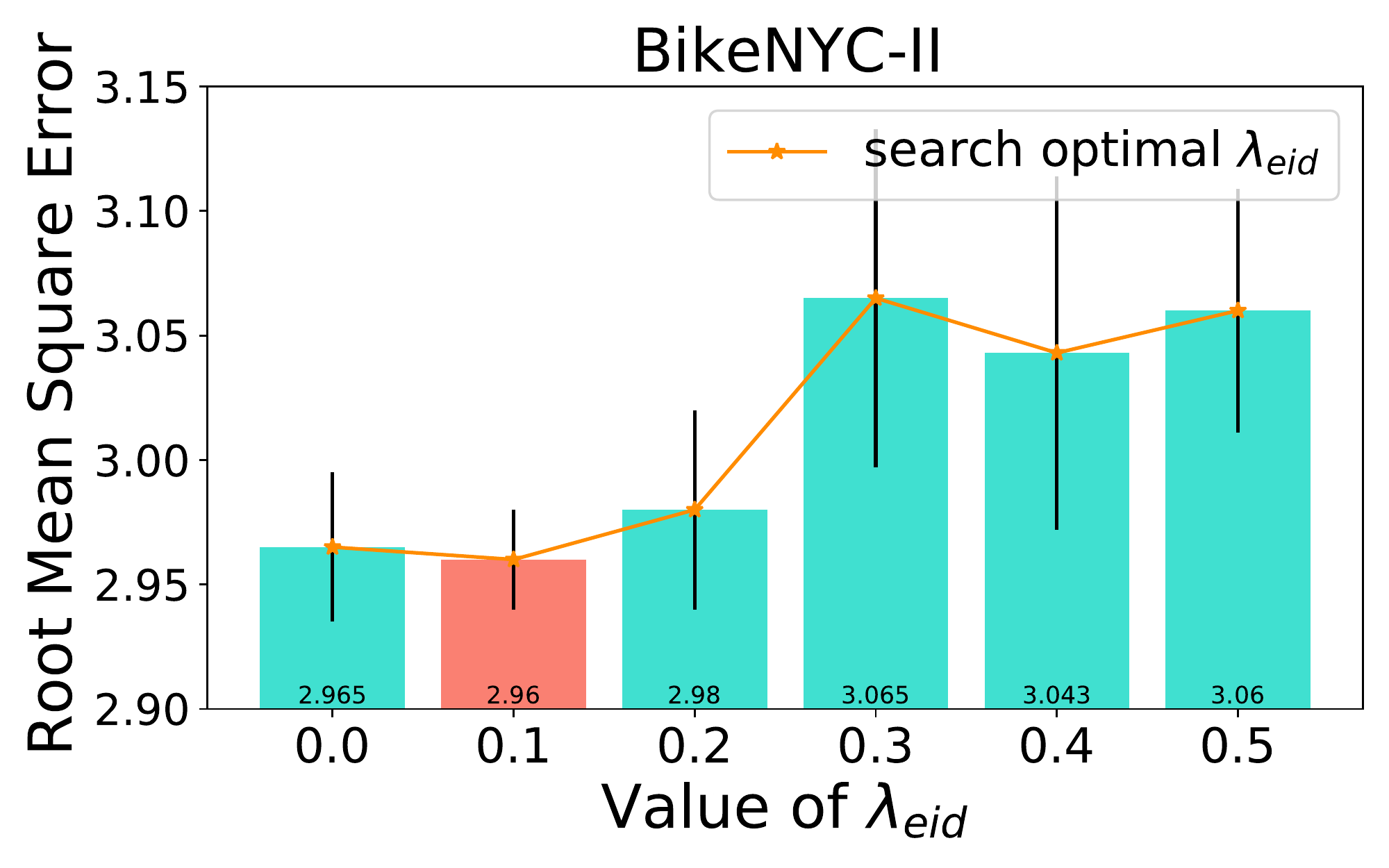}}
	\hfill
	\subfloat{\includegraphics[width=0.24\textwidth]{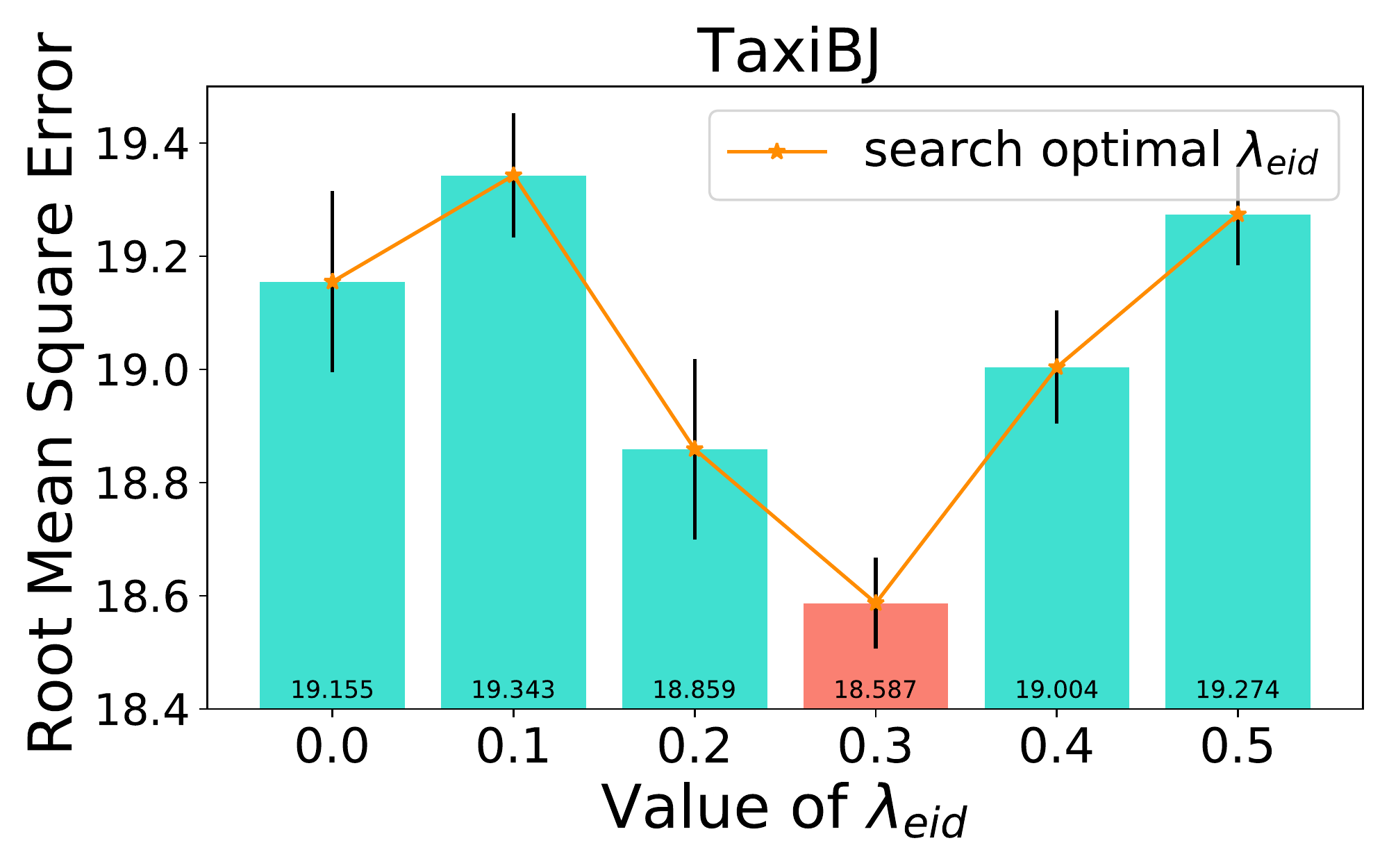}}
	\hfill
	\subfloat{\includegraphics[width=0.24\textwidth]{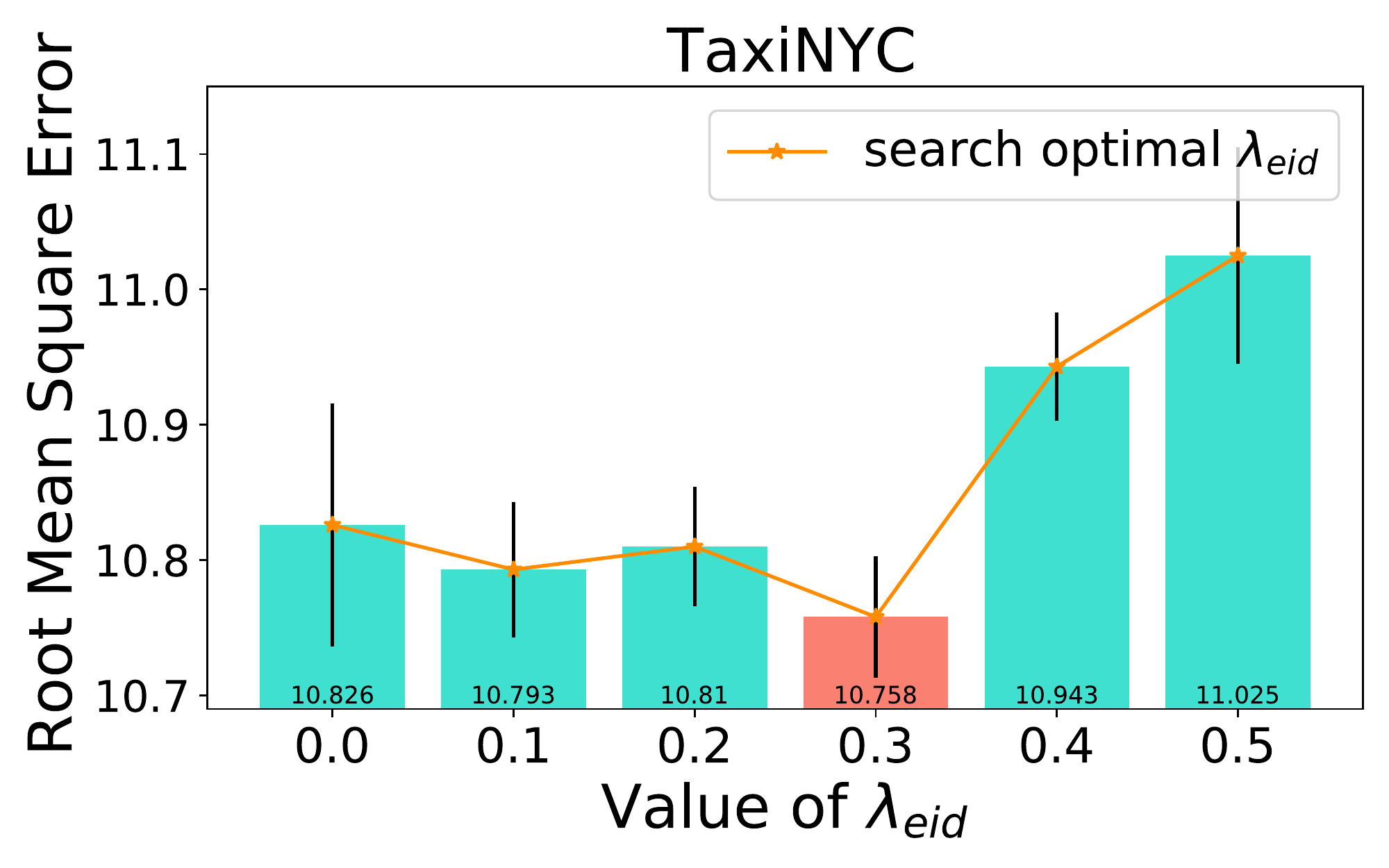}}
	\caption{Parameter analysis. }\label{fig:Parameter_analysis}
\end{figure*}

\subsection{Result Analysis}
We first compare with seven other models in TaxiBJ, TaxiNYC, BikeNYC-I, and BikeNYC-II, as shown in \tableref{tab:result1} and \tableref{tab:result2}. We give three variants of ST-ExpertNet by adopting different architectures: CNN, ConvLSTM and ST-ResNet, which are often deployed in crowd flow prediction task. 
The results in \tableref{tab:result1}  and  \tableref{tab:result2} shows that these baseline models have been greatly improved by adapting them into ST-ExpertNet. By observing experimentally, we find that for weak learners, the network is suitable to set up more experts, and each weak learner can learn a basic pattern from input, which may be easier to train and achieve higher improvements than other complicated learners. Further, we discover that data sets will also have influences on different architectures of ST-ExpertNet. For example, CNN-Expert-Net gets 137\%, 144\%, and 304\% improved in NYC, but still has a big gap compared to the state of the arts in TaxiBJ. We consider the reason is that the time span in TaxiBJ is large and the relationship between regions and patterns processes higher nonlinearity, which may be more unpredictable for weak learners. On the contrary, the TaxiNYC and BikeNYC datasets contains less than three months data, which may be suitable for simple models. For ConvLSTM and ST-ResNet in ST-ExpertNet, we see that the model still gets satisfying
performance with only a few number of experts. For more detailed information on model running time and parameter scales, please refer to supplementary materials.



\subsection{Parameter Analysis}

	\noindent\textbf{Statistical Analysis} To statistically analyze the discrepancy between experts, the Quade test \cite{quade1979using} is used, which is a non-parametric test to see if $k$ experiment outcomes have identical effects. The null hypothesis of the test assumes an identical distribution for all test cases. In our paper, we set $k$=2 and compute the pairwise p-value for all the experts. Formally, given two expert outputs $Y_1, Y_2 \in \mathbb{R}^{n\times 2 \times h \times w}$, where $n$ denotes the number of samples (batch size), we calculate the p-value between two experts as
	\begin{align*}
		p-value=\frac{1}{2\times h\times w}\sum_{c=1}^{2}\sum_{i=0}^{h}\sum_{j=0}^{w} Quade(Y_1^{:,c,h,w},Y_2^{:,c,h,w}).
	\end{align*}
	The $Quade$ function outputs the significance level of the Quade test and $c$ denotes the index of in-out flow channel. The final p-value is calculated as the arithmetic mean of significance levels from two experts' regional outputs, and a smaller p-value indicates higher variance between the two experts. \figref{Quade Test} shows the pairwise p-values of the test results for all 10 experts in CNN-ExpertNet under $\lambda_{eid}=0$ and $\lambda_{eid}=0.1$. It is obvious that pairwise p-values are much lower after using \textit{Experts Inter Discrepancy Loss} by setting $\lambda_{eid}=0.1$. It is also noted that the max pairwise p-value under $\lambda_{eid}=0.1$ is less than 0.05, indicating it's statistically significant to show the diversity of experts.


\noindent\textbf{Ablation Study} We also conduct the ablation experiment on TaxiBJ and TaxiNYC with $G_s$ and $G_t$ being masked separately.
Removing $G_s$ and using the expert's output directly to deploy self-attention in the model can mask $G_s$. For masking $G_t$, simply throwing away $G_t$ will be fine. The experiment includes CNN, ResNet and ConvLSTM as base experts and the result is shown in \tableref{tab:ablation_result} with comparison to the original models.
We can observe that $G_s$ and $G_t$ all play a crucial role in both datasets and improve performance significantly. 

\begin{table}[t]
	\footnotesize
	\centering
	\caption{ Ablation study on TaxiBJ and TaxiNYC. ($\neg$ denotes the mask operation)  }\label{tab:ablation_result}
	\begin{tabular}{|l|c|c|c|c|c|}
		\hline & \multicolumn{2}{|c|} { \textbf{TaxiBJ} } & \multicolumn{2}{|c|} { \textbf{TaxiNYC} } \\
		\hline Model & MSE  & MAPE & MSE  & MAPE  \\
		CNN-ExpertNet &{342.737} &  {5.28}\%& 114.717  &     4.50\%\\
		CNN-ExpertNet$\neg G_s$  &470.824&   6.58\% &131.002& 5.15\%\\
		CNN-ExpertNet$\neg G_t$ &444.237& 6.23\% &153.528 & 5.20\% \\
		
		ST-ResNet-ExpertNet &330.258& 5.05\% &111.445 & 4.04\% \\
		
		ST-ResNet-ExpertNet$\neg G_s$ &357.201& 5.34\% &126.173 & 4.17\% \\
		
		ST-ResNet-ExpertNet$\neg G_t$ &343.534& 5.28\% &155.113 & 5.33\% \\
		
		ConvLSTM-ExpertNet &\textbf{328.250} & \textbf{4.95}\% &\textbf{109.445} & \textbf{3.93}\% \\
		
		ConvLSTM-ExpertNet$\neg G_s$ &344.776& 5.87\% &110.318 & 4.24\% \\
		
		ConvLSTM-ExpertNet$\neg G_t$ &331.034& 5.16\% &117.772 & 4.62\% \\
		\hline
	\end{tabular}
\end{table}
\noindent\textbf{Parameters Search}.  
We examine the sensitivities of three important hyperparameters: the number of experts $K$, the punishment intensity of $\lambda_{er}$, the punishment intensity of $\lambda_{eid}$ in CNN-ExpertNet training on four datasets. Each experiment will repeat three times, and the mean and variance has been calculated, and depicted in  \figref{fig:Parameter_analysis} as the format of bar plot. The best results are marked as red color. For the sake of fairness, the experiment of searching optimal $K$ will firstly run, then, the optimal $K$ will be set as the default number of expert in the experiment of $\lambda_{er}$ and $\lambda_{eid}$. For briefly , we summarize several conclusions from \figref{fig:Parameter_analysis}:
\begin{itemize}
	\item $K$: The number of experts corresponds to the complexity of the data set. For example, we can discover that the order of estimation error for all baselines in \tableref{tab:result1} and \ref{tab:result2} is TaxiBJ $>$ TaxiNYC $>$ BikeNYC-I $>$ BikeNYC-II, and the optimal $K$ are 9, 7, 4, 2 respectively. Therefore, when the performance of ST-ExpertNet is inferior to such as ST-ResNet, or ConvLSTM,
	slightly enlarging the experts is encouraged. 
	\item $\lambda_{er}:$ From \figref{fig:Parameter_analysis} known, RMSE is not sensitive to $\lambda_{er}$ in contrast to $\lambda_{eid}$. The function of $\lambda_{er}$ is to push the  coordinated between the output of expert and spatial gating, and enhance the interpretation of ST-ExpertNet, which means that the irrelevant expert's output can be vanished through $\lambda_{er}$.   
	
	\item $\lambda_{eid}:$ Intuitively, the more experts, the more entanglement/overlap between experts. Therefore, we can observe that in TaxiBJ  TaxiNYC, BikeNYC-I,  BikeNYC-II,  the optimal $K$ are 9, 7, 4, 2 and the optimal $\lambda_{eid}$ are 0.3, 0.3, 0.1, 0.1, which proves that $\lambda_{eid}$ corresponds to $K$.
\end{itemize}

\begin{figure}[t]
	\centering
	\includegraphics[width=1\linewidth]{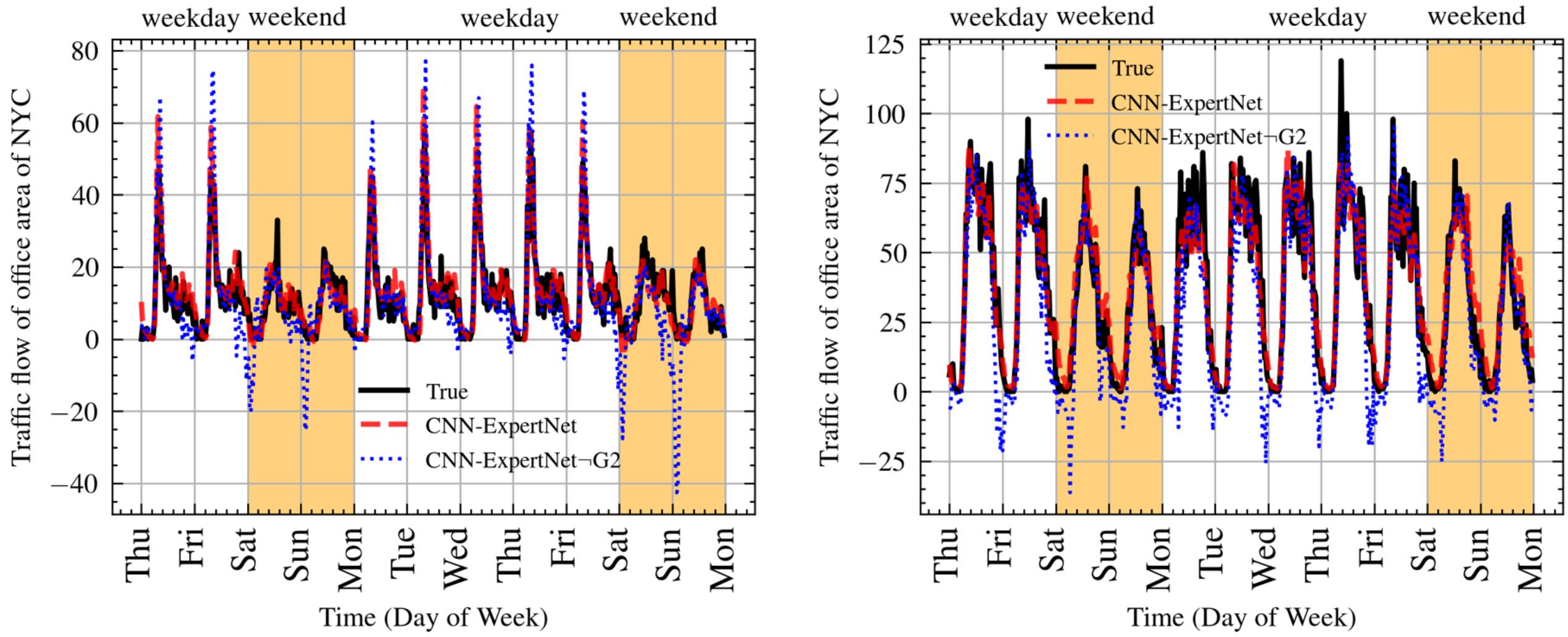}
	\caption{The performance of CNN-ExpertNet and CNN-ExpertNet$\neg$G2 (CNN-ExpertNet without $G_t$) on TaxiNYC dataset.}
	\label{fig:without_g2}
	\vspace{-0.4cm}
\end{figure}

\subsection{Case Study with Gating Network $G_s$ in TaxiBJ}
We evaluate $G_s$ 's spatial self-adaption and interpretability by visualizing the attention values for each expert $E_i$. CNN-ExpertNet's inflow spatial attention values at 8:00 am are computed. The top six experts' attention values are visualized and shown on \figref{fig:attentionspatial}. We can observe that the attention of each expert is varied. For instance,  expert 5 concentrates on the commuting pattern of Beijing's 2nd, 3rd, and 4th ring roads, China's first fully enclosed, full interchange, no-traffic-light urban express ring road. Compared to expert 5, expert 2 is responsible for the 4th and 5th ring roads in Beijing. Furthermore, expert 3 takes charge of the residential area in the Haidian and Fengtai districts, which  
contain multiple kinds of places of interest, such as Summer Palace, Xiangshan Park, Beijing Garden Expo Park, etc. Expert 4 mainly charges the Capital Airport Expressway, an expressway connecting downtown Beijing and Beijing Capital International Airport, and contains huge flow values throughout the day. Since multiple patterns may exist in one region simultaneously, expert 1 co-manages the flow state with experts 2 and 5 in some places. Expert 6 mainly controls the suburban area in the southwest area of Beijing, where many villages and town exists, and the model considers 
this area flow pattern is different from others. Therefore, we can conclude that ST-ExpertNet successfully learns the different flow patterns in our city.


We also select two region types from TaxiBJ; one is the office area, and the other is the residential area. There are two conspicuous patterns for the office and residential areas. The inflow pattern for the previous one has morning and evening peaks every day. On the contrary, people often leave their homes and go to work on workdays. Therefore, the outflow of residential in peaking hours tends to be higher than another timestamp. The \figref{fig:attentiontemporal} A and B present the change of flows at the coordinate of (2,31)  and (11,23) on 5/20/2015, respectively. It can be seen that in different periods, the regional traffic flow is dominated by a certain expert
\figref{fig:attentiontemporal} C
presents the office area attention value. We can observe that the duty of $E_5$ is mainly to adjust the flow distribution for the midnight and the morning and evening peak flow output is mainly managed by $E_8$. In \figref{fig:attentiontemporal} D, the residential area pattern is mainly charged by $E_{1}$ and $E_{3}$. The $E_{1}$ controls the midnight pattern, and the $E_{3}$ pays attention to the morning peak pattern.


\subsection{Case Study with Gating Network $G_t$ in TaxiNYC}
We investigate whether $G_t$ can be extended to learn meaningful temporal patterns and modify the amplitude of time series output in the real-world large scalar dataset. Therefore, the ablation study has been conducted on the TaxiNYC dataset to demonstrate that $G_t$ helps adjust the expert temporal signal. \figref{fig:without_g2} shows an example of the predictions with ST-ExpertNet with $G_t$ and  without $G_t$. The ground truth data of the figure is from the inflow of TaxiNYC at coordinate (0,2), where an office area locates. It can be viewed from the graph that on weekdays the inflow peaks in the morning and approaches nearly zero at night, whereas on weekends, the inflow has a much lower peak in the morning and reaches almost zero in the evening. However, we can see from the result that the network without $G_t$ tends to have unstable predictions, especially in the morning and at late night. Besides, it can't handle the temporal information well because every weekend, when there is a sudden shift of predictive pattern from weekdays to weekends, it will have trouble dealing with the late-night prediction. As a comparison, the
predictive result of the network with $G_t$ is much more satisfying. The Gating Network $G_t$ learns the temporal pattern and helps the network adjust its output based on the input data, which consists of historical information. When it comes to weekends, $G_t$ will help the network switch to the ``predictive weekend pattern'' to gain a more accurate performance.

%
%


%
\section{Conclusion}\label{sec:concluds}
This paper introduced a new deep learning-based framework for city-wide crowd flow prediction using several experts based on historical spatial-temporal data, weather, and holiday events. The ST-ExpertNet can decompose the crowd flow tensor into several patterns (e.g., working, entertaining, commuting, commercial, etc.) and localizes different experts into different patterns. The final output of the framework is a weighted sum of all the experts via the attention mechanism, which can be visualized and, to some extent, increases the framework's
interpretability. Furthermore, experts in our framework can be set as any suitable model by users, such as CNN, ResNet, and ConvLSTM.
We comprehensively investigate our framework's performance on four types of crowd flows in Beijing and NYC compared to the other seven baseline methods. The experiment result demonstrates that ST-ExpertNet can achieve better predictive accuracy on these datasets and is more applicable to crowd flow prediction.

As a limitation of this study, the state-of-the-art is
hard to directly employ into the ST-ExpertNet 
framework due to their complex architecture.
On the other hand, straightforward models are encouraged since ST-ExpertNet is attributable to the ensemble framework.
In the future, experts can be tried to build different functions, such as dividing the day according to specific rules (e.g., holidays, workday, weekends), since people's traffic patterns range from day to day. Another direction can be considered the traffic flow as a time series with trends, seasonality, and residual components. Each expert can learn these components as well as the time, date, and region.

\section*{Acknowledgment}
Acknowledgment: This work was partially supported by National Key Research and Development Project (2021YFB1714400) of China and  Guangdong Provincial Key Laboratory (2020B121201001).

\bibliographystyle{IEEEtran}
\bibliography{citation}

\begin{thebibliography}{10}
\providecommand{\url}[1]{#1}
\csname url@samestyle\endcsname
\providecommand{\newblock}{\relax}
\providecommand{\bibinfo}[2]{#2}
\providecommand{\BIBentrySTDinterwordspacing}{\spaceskip=0pt\relax}
\providecommand{\BIBentryALTinterwordstretchfactor}{4}
\providecommand{\BIBentryALTinterwordspacing}{\spaceskip=\fontdimen2\font plus
\BIBentryALTinterwordstretchfactor\fontdimen3\font minus
  \fontdimen4\font\relax}
\providecommand{\BIBforeignlanguage}[2]{{%
\expandafter\ifx\csname l@#1\endcsname\relax
\typeout{** WARNING: IEEEtran.bst: No hyphenation pattern has been}%
\typeout{** loaded for the language `#1'. Using the pattern for}%
\typeout{** the default language instead.}%
\else
\language=\csname l@#1\endcsname
\fi
#2}}
\providecommand{\BIBdecl}{\relax}
\BIBdecl

\bibitem{zhang2017deep}
J.~Zhang, Y.~Zheng, and D.~Qi, ``Deep spatio-temporal residual networks for
  citywide crowd flows prediction,'' in \emph{Thirty-First AAAI Conference on
  Artificial Intelligence}, 2017.

\bibitem{yao2019revisiting}
H.~Yao, X.~Tang, H.~Wei, G.~Zheng, and Z.~Li, ``Revisiting spatial-temporal
  similarity: A deep learning framework for traffic prediction,'' in
  \emph{Proceedings of the AAAI conference on artificial intelligence},
  vol.~33, no.~01, 2019, pp. 5668--5675.

\bibitem{lin2019deepstn}
Z.~Lin, J.~Feng, Z.~Lu, Y.~Li, and D.~Jin, ``Deepstn+: Context-aware
  spatial-temporal neural network for crowd flow prediction in metropolis,'' in
  \emph{Proceedings of the AAAI Conference on Artificialsimonyan2013deep
  Intelligence}, vol.~33, 2019, pp. 1020--1027.

\bibitem{zonoozi2018periodic}
A.~Zonoozi, J.-j. Kim, X.-L. Li, and G.~Cong, ``Periodic-crn: A convolutional
  recurrent model for crowd density prediction with recurring periodic
  patterns.'' in \emph{IJCAI}, 2018, pp. 3732--3738.

\bibitem{yuan2012discovering}
J.~Yuan, Y.~Zheng, and X.~Xie, ``Discovering regions of different functions in
  a city using human mobility and pois,'' in \emph{Proceedings of the 18th ACM
  SIGKDD international conference on Knowledge discovery and data mining},
  2012, pp. 186--194.

\bibitem{yin2011geographical}
Z.~Yin, L.~Cao, J.~Han, C.~Zhai, and T.~Huang, ``Geographical topic discovery
  and comparison,'' in \emph{Proceedings of the 20th international conference
  on World wide web}, 2011, pp. 247--256.

\bibitem{wang2014discovering}
J.~Wang, F.~Gao, P.~Cui, C.~Li, and Z.~Xiong, ``Discovering urban
  spatio-temporal structure from time-evolving traffic networks,'' in
  \emph{Asia-Pacific Web Conference}.\hskip 1em plus 0.5em minus 0.4em\relax
  Springer, 2014, pp. 93--104.

\bibitem{fan2014cityspectrum}
Z.~Fan, X.~Song, and R.~Shibasaki, ``Cityspectrum: a non-negative tensor
  factorization approach,'' in \emph{Proceedings of the 2014 ACM International
  Joint Conference on Pervasive and Ubiquitous Computing}, 2014, pp. 213--223.

\bibitem{jacobs1991task}
R.~A. Jacobs, M.~I. Jordan, and A.~G. Barto, ``Task decomposition through
  competition in a modular connectionist architecture: The what and where
  vision tasks,'' \emph{Cognitive science}, vol.~15, no.~2, pp. 219--250, 1991.

\bibitem{jacobs1997bias}
R.~A. Jacobs, ``Bias/variance analyses of mixtures-of-experts architectures,''
  \emph{Neural computation}, vol.~9, no.~2, pp. 369--383, 1997.

\bibitem{jordan1994hierarchical}
M.~I. Jordan and R.~A. Jacobs, ``Hierarchical mixtures of experts and the em
  algorithm,'' \emph{Neural computation}, vol.~6, no.~2, pp. 181--214, 1994.

\bibitem{jacobs1991adaptive}
R.~A. Jacobs, M.~I. Jordan, S.~J. Nowlan, and G.~E. Hinton, ``Adaptive mixtures
  of local experts,'' \emph{Neural computation}, vol.~3, no.~1, pp. 79--87,
  1991.

\bibitem{pang2019improving}
T.~Pang, K.~Xu, C.~Du, N.~Chen, and J.~Zhu, ``Improving adversarial robustness
  via promoting ensemble diversity,'' in \emph{International Conference on
  Machine Learning}.\hskip 1em plus 0.5em minus 0.4em\relax PMLR, 2019, pp.
  4970--4979.

\bibitem{zhang2019flow}
J.~Zhang, Y.~Zheng, J.~Sun, and D.~Qi, ``Flow prediction in spatio-temporal
  networks based on multitask deep learning,'' \emph{IEEE Transactions on
  Knowledge and Data Engineering}, vol.~32, no.~3, pp. 468--478, 2019.

\bibitem{yao2018deep}
H.~Yao, F.~Wu, J.~Ke, X.~Tang, Y.~Jia, S.~Lu, P.~Gong, J.~Ye, and Z.~Li, ``Deep
  multi-view spatial-temporal network for taxi demand prediction,'' in
  \emph{Proceedings of the AAAI Conference on Artificial Intelligence},
  vol.~32, no.~1, 2018.

\bibitem{bahdanau2014neural}
D.~Bahdanau, K.~Cho, and Y.~Bengio, ``Neural machine translation by jointly
  learning to align and translate,'' \emph{arXiv preprint arXiv:1409.0473},
  2014.

\bibitem{masoudnia2014mixture}
S.~Masoudnia and R.~Ebrahimpour, ``Mixture of experts: a literature survey,''
  \emph{Artificial Intelligence Review}, vol.~42, no.~2, pp. 275--293, 2014.

\bibitem{tang2002input}
B.~Tang, M.~I. Heywood, and M.~Shepherd, ``Input partitioning to mixture of
  experts,'' in \emph{Proceedings of the 2002 International Joint Conference on
  Neural Networks. IJCNN'02 (Cat. No. 02CH37290)}, vol.~1.\hskip 1em plus 0.5em
  minus 0.4em\relax IEEE, 2002, pp. 227--232.

\bibitem{gutta2000mixture}
S.~Gutta, J.~R. Huang, P.~Jonathon, and H.~Wechsler, ``Mixture of experts for
  classification of gender, ethnic origin, and pose of human faces,''
  \emph{IEEE Transactions on neural networks}, vol.~11, no.~4, pp. 948--960,
  2000.

\bibitem{goodband2006mixture}
J.~Goodband, O.~C. Haas, and J.~A. Mills, ``A mixture of experts committee
  machine to design compensators for intensity modulated radiation therapy,''
  \emph{Pattern recognition}, vol.~39, no.~9, pp. 1704--1714, 2006.

\bibitem{hansen1999combining}
J.~V. Hansen, ``Combining predictors: comparison of five meta machine learning
  methods,'' \emph{Information Sciences}, vol. 119, no. 1-2, pp. 91--105, 1999.

\bibitem{ubeyli2010differentiation}
E.~D. {\"U}beyli, K.~Ilbay, G.~Ilbay, D.~Sahin, and G.~Akansel,
  ``Differentiation of two subtypes of adult hydrocephalus by mixture of
  experts,'' \emph{Journal of medical systems}, vol.~34, no.~3, pp. 281--290,
  2010.

\bibitem{collobert2002parallel}
R.~Collobert, S.~Bengio, and Y.~Bengio, ``A parallel mixture of svms for very
  large scale problems,'' \emph{Neural computation}, vol.~14, no.~5, pp.
  1105--1114, 2002.

\bibitem{tresp2001mixtures}
V.~Tresp, ``Mixtures of gaussian processes,'' \emph{Advances in neural
  information processing systems}, pp. 654--660, 2001.

\bibitem{shahbaba2009nonlinear}
B.~Shahbaba and R.~Neal, ``Nonlinear models using dirichlet process mixtures.''
  \emph{Journal of Machine Learning Research}, vol.~10, no.~8, 2009.

\bibitem{yao2009hierarchical}
B.~Yao, D.~Walther, D.~Beck, and L.~Fei-Fei, ``Hierarchical mixture of
  classification experts uncovers interactions between brain regions,''
  \emph{Advances in Neural Information Processing Systems}, vol.~22, pp.
  2178--2186, 2009.

\bibitem{rasmussen2002infinite}
C.~E. Rasmussen and Z.~Ghahramani, ``Infinite mixtures of gaussian process
  experts,'' \emph{Advances in neural information processing systems}, vol.~2,
  pp. 881--888, 2002.

\bibitem{aljundi2017expert}
R.~Aljundi, P.~Chakravarty, and T.~Tuytelaars, ``Expert gate: Lifelong learning
  with a network of experts,'' in \emph{Proceedings of the IEEE Conference on
  Computer Vision and Pattern Recognition}, 2017, pp. 3366--3375.

\bibitem{yildirim2019eboc}
P.~Y{\i}ld{\i}r{\i}m, U.~K. Birant, and D.~Birant, ``Eboc: Ensemble-based
  ordinal classification in transportation,'' \emph{Journal of Advanced
  Transportation}, vol. 2019, 2019.

\bibitem{bangui2021recent}
H.~Bangui and B.~Buhnova, ``Recent advances in machine-learning driven
  intrusion detection in transportation: Survey,'' \emph{Procedia Computer
  Science}, vol. 184, pp. 877--886, 2021.

\bibitem{wen2015rapid}
X.~Wen, L.~Shao, Y.~Xue, and W.~Fang, ``A rapid learning algorithm for vehicle
  classification,'' \emph{Information sciences}, vol. 295, pp. 395--406, 2015.

\bibitem{deri2016big}
J.~A. Deri, F.~Franchetti, and J.~M. Moura, ``Big data computation of taxi
  movement in new york city,'' in \emph{2016 IEEE International Conference on
  Big Data (Big Data)}.\hskip 1em plus 0.5em minus 0.4em\relax IEEE, 2016, pp.
  2616--2625.

\bibitem{zhan2016citywide}
X.~Zhan, Y.~Zheng, X.~Yi, and S.~V. Ukkusuri, ``Citywide traffic volume
  estimation using trajectory data,'' \emph{IEEE Transactions on Knowledge and
  Data Engineering}, vol.~29, no.~2, pp. 272--285, 2016.

\bibitem{said1984testing}
S.~E. Said and D.~A. Dickey, ``Testing for unit roots in autoregressive-moving
  average models of unknown order,'' \emph{Biometrika}, vol.~71, no.~3, pp.
  599--607, 1984.

\bibitem{cheng2017traffic}
Y.~Cheng, X.~Cheng, M.~Tan, K.~ZHOU, and H.-b. LI, ``Traffic flow prediction
  based on hybrid model of arima and wnn,'' \emph{Computer Technology \&
  Development}, vol.~27, no.~1, pp. 169--172, 2017.

\bibitem{box1970distribution}
G.~E. Box and D.~A. Pierce, ``Distribution of residual autocorrelations in
  autoregressive-integrated moving average time series models,'' \emph{Journal
  of the American statistical Association}, vol.~65, no. 332, pp. 1509--1526,
  1970.

\bibitem{zhang2016dnn}
J.~Zhang, Y.~Zheng, D.~Qi, R.~Li, and X.~Yi, ``Dnn-based prediction model for
  spatio-temporal data,'' in \emph{Proceedings of the 24th ACM SIGSPATIAL
  International Conference on Advances in Geographic Information Systems},
  2016, pp. 1--4.

\bibitem{he2016deep}
K.~He, X.~Zhang, S.~Ren, and J.~Sun, ``Deep residual learning for image
  recognition,'' in \emph{Proceedings of the IEEE conference on computer vision
  and pattern recognition}, 2016, pp. 770--778.

\bibitem{wang2020multi}
S.~Wang, H.~Miao, H.~Chen, and Z.~Huang, ``Multi-task adversarial
  spatial-temporal networks for crowd flow prediction,'' in \emph{Proceedings
  of the 29th ACM international conference on information \& knowledge
  management}, 2020, pp. 1555--1564.

\bibitem{du2019deep}
B.~Du, H.~Peng, S.~Wang, M.~Z.~A. Bhuiyan, L.~Wang, Q.~Gong, L.~Liu, and J.~Li,
  ``Deep irregular convolutional residual lstm for urban traffic passenger
  flows prediction,'' \emph{IEEE Transactions on Intelligent Transportation
  Systems}, vol.~21, no.~3, pp. 972--985, 2019.

\bibitem{zhao2020discover}
X.~Zhao, Z.~Li, Y.~Zhang, and Y.~Lv, ``Discover trip purposes from cellular
  network data with topic modelling,'' \emph{IEEE Intelligent Transportation
  Systems Magazine}, 2020.

\bibitem{pozdnoukhov2011space}
A.~Pozdnoukhov and C.~Kaiser, ``Space-time dynamics of topics in streaming
  text,'' in \emph{Proceedings of the 3rd ACM SIGSPATIAL international workshop
  on location-based social networks}, 2011, pp. 1--8.

\bibitem{sun2016understanding}
L.~Sun and K.~W. Axhausen, ``Understanding urban mobility patterns with a
  probabilistic tensor factorization framework,'' \emph{Transportation Research
  Part B: Methodological}, vol.~91, pp. 511--524, 2016.

\bibitem{oord2016conditional}
A.~v.~d. Oord, N.~Kalchbrenner, O.~Vinyals, L.~Espeholt, A.~Graves, and
  K.~Kavukcuoglu, ``Conditional image generation with pixelcnn decoders,''
  \emph{arXiv preprint arXiv:1606.05328}, 2016.

\bibitem{krogh1995validation}
A.~Krogh and J.~Vedelsby, ``Validation, and active learning,'' \emph{Advances
  in neural information processing systems 7}, vol.~7, p. 231, 1995.

\bibitem{kulesza2012determinantal}
A.~Kulesza and B.~Taskar, ``Determinantal point processes for machine
  learning,'' \emph{arXiv preprint arXiv:1207.6083}, 2012.

\bibitem{Matrixmathematics}
D.~S. Bernstein, \emph{Matrix mathematics: Theory, facts, and formulas with
  application to linear systems theory}.\hskip 1em plus 0.5em minus 0.4em\relax
  Princeton university press Princeton, 2005.

\bibitem{jiang2021dl}
R.~Jiang, D.~Yin, Z.~Wang, Y.~Wang, J.~Deng, H.~Liu, Z.~Cai, J.~Deng, X.~Song,
  and R.~Shibasaki, ``Dl-traff: Survey and benchmark of deep learning models
  for urban traffic prediction,'' in \emph{Proceedings of the 30th ACM
  International Conference on Information \& Knowledge Management}, 2021, pp.
  4515--4525.

\bibitem{shi2015convolutional}
X.~Shi, Z.~Chen, H.~Wang, D.-Y. Yeung, W.-K. Wong, and W.-c. Woo,
  ``Convolutional lstm network: A machine learning approach for precipitation
  nowcasting,'' \emph{arXiv preprint arXiv:1506.04214}, 2015.

\bibitem{zhao2019long}
J.~Zhao, F.~Deng, Y.~Cai, and J.~Chen, ``Long short-term memory-fully connected
  (lstm-fc) neural network for pm2. 5 concentration prediction,''
  \emph{Chemosphere}, vol. 220, pp. 486--492, 2019.

\bibitem{ballas2015delving}
N.~Ballas, L.~Yao, C.~Pal, and A.~Courville, ``Delving deeper into
  convolutional networks for learning video representations,'' \emph{arXiv
  preprint arXiv:1511.06432}, 2015.

\bibitem{paszke2019pytorch}
A.~Paszke, S.~Gross, F.~Massa, A.~Lerer, J.~Bradbury, G.~Chanan, T.~Killeen,
  Z.~Lin, N.~Gimelshein, L.~Antiga \emph{et~al.}, ``Pytorch: An imperative
  style, high-performance deep learning library,'' \emph{arXiv preprint
  arXiv:1912.01703}, 2019.

\bibitem{kingma2014adam}
D.~P. Kingma and J.~Ba, ``Adam: A method for stochastic optimization,''
  \emph{arXiv preprint arXiv:1412.6980}, 2014.

\bibitem{quade1979using}
D.~Quade, ``Using weighted rankings in the analysis of complete blocks with
  additive block effects,'' \emph{Journal of the American Statistical
  Association}, vol.~74, no. 367, pp. 680--683, 1979.

\end{thebibliography}
\begin{IEEEbiography}[{\includegraphics[width=1in,height=1.25in,clip,keepaspectratio]{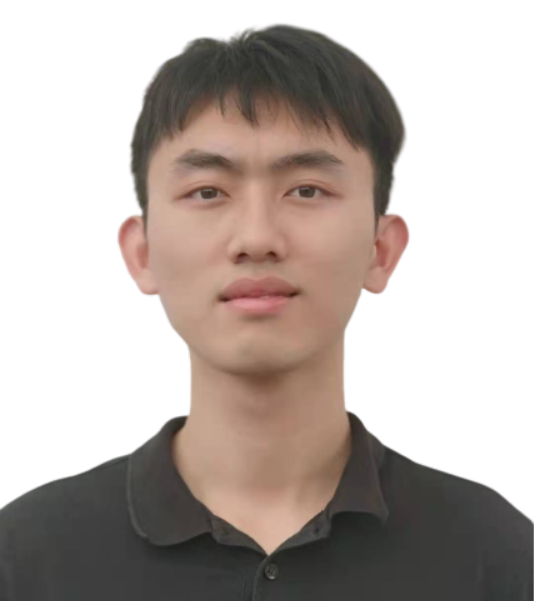}}]{Hongjun Wang} is working toward the M.S. degree
	in  computer science and technology from Southern University of Science and Technology, China. He received the B.E. degree from the Nanjing University of Posts and Telecommunications, China, in 2019. His research interests
	are broadly in machine learning, with urban computing, explainable AI, data mining, data visualization.
\end{IEEEbiography}
\vspace{-11ex}
\begin{IEEEbiography}[{\includegraphics[width=1in,height=1.25in,clip,keepaspectratio]{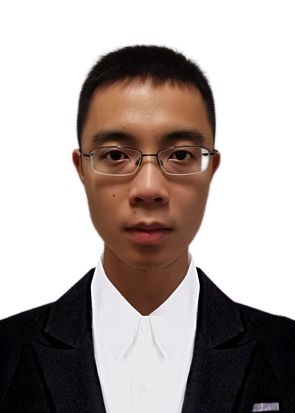}}]{Jiyuan Chen} is working towards his B.S. degree in Computer Science and Technology from Southern University of Science and Technology, China. His major research fields include artificial intelligence, deep learning, urban computing and data mining. 
\end{IEEEbiography}
\vspace{-11ex}
\begin{IEEEbiography}[{\includegraphics[width=1in,height=1.25in,clip,keepaspectratio]{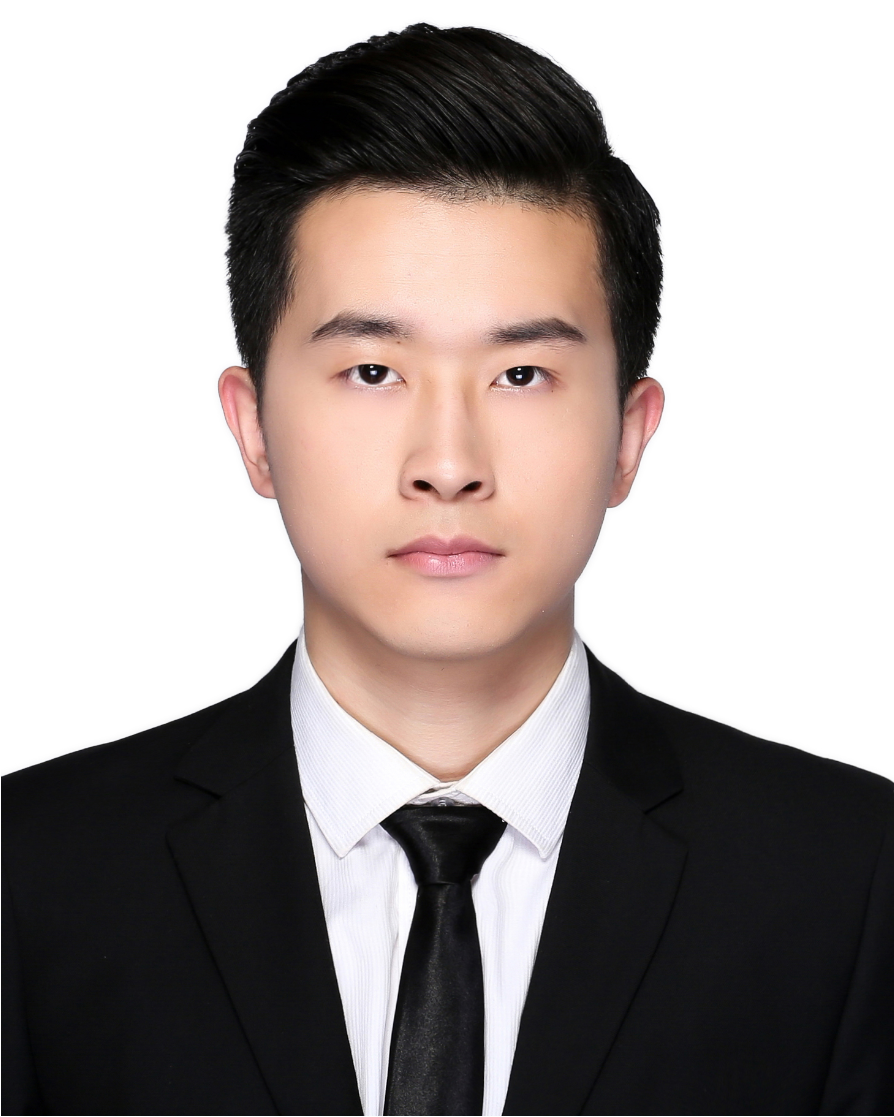}}]{Zekun Cai} received his B.S. degree in Computer Science and Technology from the University of Electronic Science and Technology of China, in 2018. From 2019, he became a master student at the Department of Socio-Cultural Environmental Studies, The University of Tokyo. His research interests are mainly focused on artificial intelligence, deep learning, ubiquitous computing, and spatio-temporal data analysis.
\end{IEEEbiography}
\vspace{-11ex}
\begin{IEEEbiography}[{\includegraphics[width=1in,height=1.25in,clip,keepaspectratio]{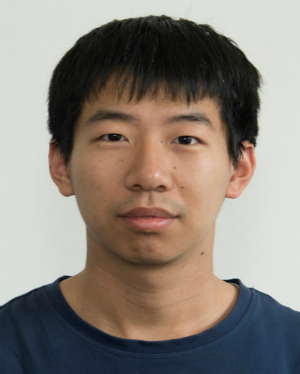}}]{Zhiwen Zhang}  received the B.E. and M.S. degree in Artificial Intelligence from Nankai University, China, in 2016 and 2019 respectively. From 2019, he is currently pursing the Ph.D. degree at Department of Socio-Cultural Environmental Studies, The University of Tokyo. His current research interests include urban computing and data visualization.
\end{IEEEbiography}
\vspace{-11ex}
\begin{IEEEbiography}[{\includegraphics[width=1in,height=1.25in,clip,keepaspectratio]{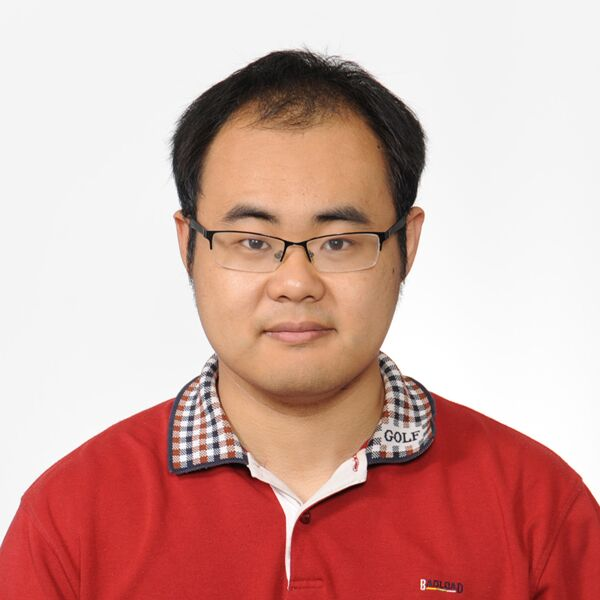}}]{Zipei Fan}  received his B.S. degree in Computer Science from Beihang University, China, in 2012, both M.S. and Ph.D. degree in Civil Engineering from The University of Tokyo, Japan, in 2014 and 2017 respectively. He became Project Researcher and Project Assistant Professor in 2017 and 2019, and he has promoted to Project Lecturer at the Center for Spatial Information Science, the University of Tokyo in 2020. His research interests include ubiquitous computing, machine learning, spatio-temporal data mining, and heterogeneous data fusion.
\end{IEEEbiography}
\vspace{-11ex}
\begin{IEEEbiography}[{\includegraphics[width=1in,height=1.25in,clip,keepaspectratio]{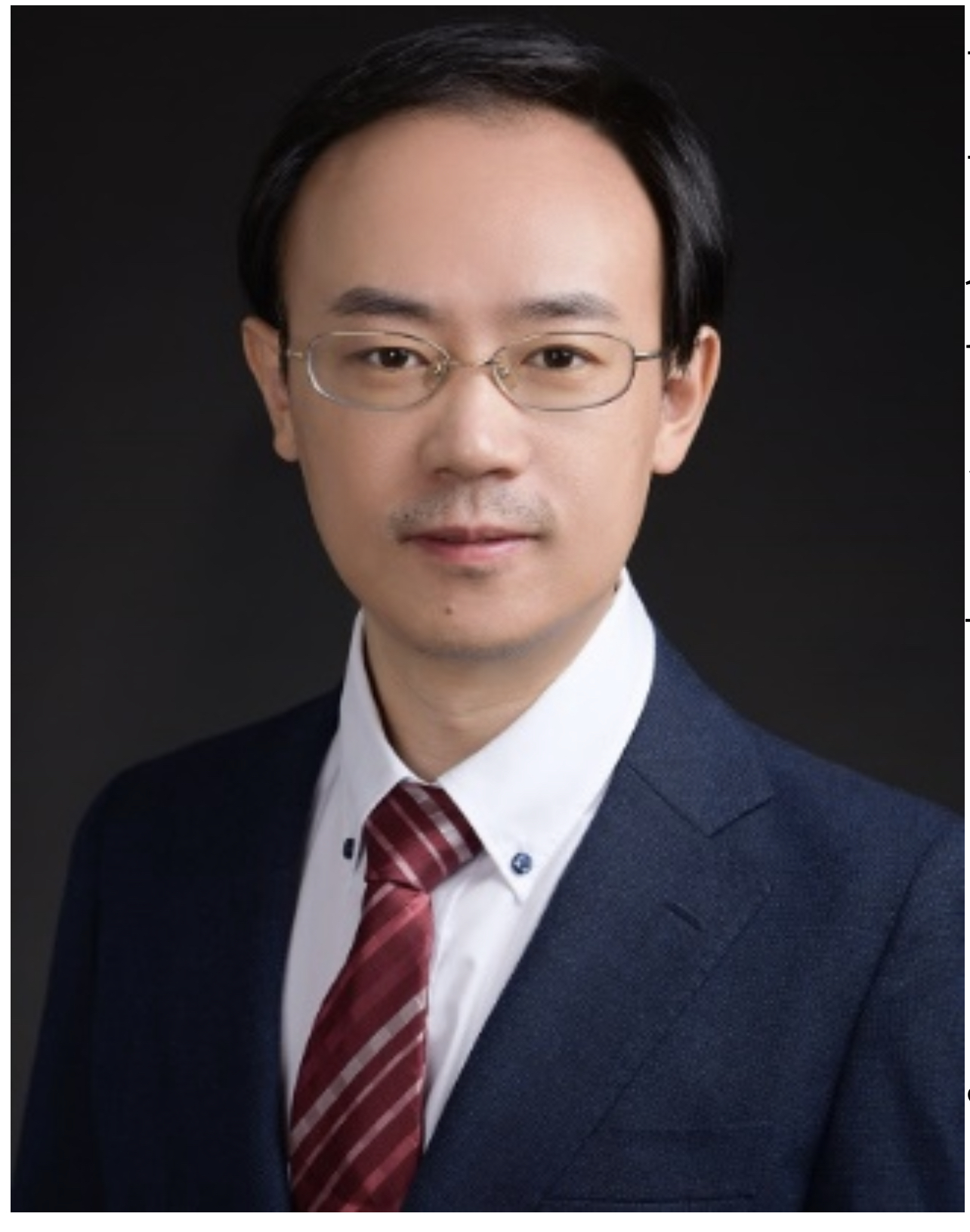}}]{Prof. Xuan Song}  received the Ph.D. degree in signal and information processing from Peking
	University in 2010. In 2017, he was selected as Excellent Young Researcher of Japan MEXT. In
	the past ten years, he led and participated in many important projects as principal investigator
	or primary actor in Japan and China. His main research interests are artificial intelligence, data science and their related fields, and he served as Associate Editor, Guest Editor, Area Chair, Program Committee Member or reviewer for many famous journals and top-tier conferences.
\end{IEEEbiography}
\vspace{-10ex}

\end{document}